\documentclass[journal]{IEEEtran}

\usepackage[utf8]{inputenc}
\usepackage[T1]{fontenc}
\usepackage{amsmath,amssymb,amsfonts}
\usepackage{mathtools}
\usepackage{graphicx}
\usepackage{subcaption}
\usepackage{cite}
\usepackage{tikz}
\usetikzlibrary{arrows.meta, shapes.geometric, positioning}
\usepackage{booktabs}
\usepackage{hyperref}
\usepackage{physics}
\usepackage{multirow}
\usepackage[ruled]{algorithm2e}
\SetKwProg{Function}{Function}{:}{end}
\usepackage{siunitx}
\AtBeginDocument{\RenewCommandCopy\qty\SI}

\usepackage{pgfplots}
\usepgfplotslibrary{groupplots}
\pgfplotsset{compat=1.18}

\usepackage[table]{xcolor}
\usepackage{colortbl}
\usepackage{tabularx}
\usepackage{threeparttable}

\definecolor{lightgreen}{RGB}{200, 255, 200}  
\definecolor{darkgreen}{RGB}{0, 128, 0}      
\newcommand{\greencell}[1]{\cellcolor{lightgreen}\textcolor{darkgreen}{#1}}

\definecolor{lightred}{RGB}{255, 200, 200}   
\definecolor{darkred}{RGB}{150, 0, 0}       
\newcommand{\redcell}[1]{\cellcolor{lightred}\textcolor{darkred}{#1}}

\DeclareMathOperator*{\argmax}{arg\,max}
\newcommand{\R}{\mathbb{R}}

\renewcommand{\vector}[1]{\mathbf{#1}}
\newcommand{\img}[1]{\mathbf{I}_{#1}}
\newcommand{\pc}[1]{\mathcal{P}_{#1}}
\newcommand{\flow}[1]{\mathcal{F}_{#1}}
\newcommand{\glob}{\mathrm{W}}
\newcommand{\ego}{\mathrm{E}}
\newcommand{\cc}{\mathcal{C}}
\newcommand{\cluster}[1]{\mathcal{G}^{#1}}

\newcommand{\egotf}{\mathbf{T}^{\ego}}   
\newcommand{\rot}{\mathbf{R}}
\newcommand{\trans}{\mathbf{t}}

\DeclareMathOperator{\arctantwo}{atan2}

\title{CorrelationFlow: A Training-Free Geometric Approach for LiDAR Scene Flow Estimation}

\author{
Minh-Quan Dao$^{1}$,
Yancong Lin$^{2}$,
Shiming Wang$^{3}$,
Julie Stephany Berrio Perez$^{4}$,
Holger Caesar$^{3}$
\thanks{M-Q. Dao is with Nantes Université, France (email: minhquan.dao@univ-nantes.fr).}
\thanks{Y. Lin is with the University of Nottingham, UK (email: Yancong.Lin@nottingham.ac.uk).}%
\thanks{S. Wang is with TU Delft, Netherlands (email: S.Wang-15@tudelft.nl).}%
\thanks{J.S. Berrio Perez is with the University of Queensland, Australia (email: s.berrioperez@uq.edu.au).}%
\thanks{H. Caesar is with TU Delft, Netherlands (email: h.caesar@tudelft.nl).}%

\small{
$^{1}$Nantes Université, LS2N\hspace{4pt}
$^{2}$University of Nottingham\hspace{4pt}
$^{3}$TU Delft\hspace{4pt}
$^{4}$University of Queensland\hspace{4pt}
}
}

\begin{document}

\twocolumn[{%
 \renewcommand\twocolumn[1][]{#1}
 \maketitle

 \vspace{-16px}
 
\begin{center}
    \captionsetup{type=figure}
    \includegraphics[width=\linewidth]{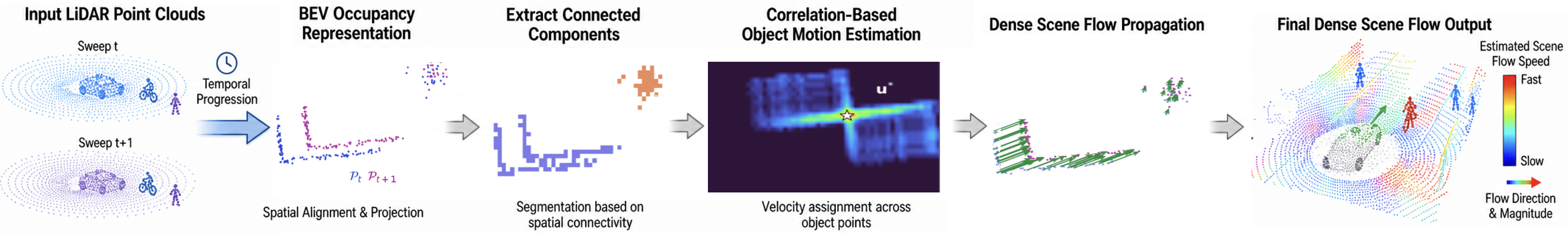}
    \captionof{figure}{\small   Overview of CorrelationFlow. Two consecutive LiDAR sweeps are projected to a BEV occupancy image, objects are isolated as spatio-temporal connected components, and each component's motion is recovered as the translation that maximizes the normalized cross-correlation of its BEV footprint across time. The per-component velocity is propagated to all member points, yielding a dense scene flow field color-coded by speed and direction.
  }
    \label{fig:pipeline}
\end{center}
 }]


\begin{abstract}
LiDAR scene flow is the task of estimating point-wise motion from sequential point clouds.
It is essential for autonomous driving, as it provides critical information about moving objects and their trajectories.
In recent years, unsupervised methods have gained significant traction by eliminating the need for expensive point-wise annotations.
However, despite incremental performance improvements, these methods uniformly rely on a single paradigm: training feed-forward networks through optimizing a loss function based on the Chamfer distance, a framework that inherently limits their progression.
This paper presents CorrelationFlow, a new paradigm that reformulates LiDAR scene flow as a correlation maximization problem between Bird's-Eye View (BEV) images.
Instead of using learnable parameters, our approach employs a chain of image processing operations to solve this problem.
The learning-free nature of CorrelationFlow enables strong generalization across diverse domains.
This capability is demonstrated by the second-place ranking in the unsupervised track of the Argoverse 2 Scene Flow Challenge at CVPR 2026, which featured a test set spanning five datasets that cover varied vehicle platforms and environments.
Our results show that this classical, model-based approach can match or even surpass the performance of learning-based methods, thereby opening a new direction for unsupervised LiDAR scene flow research.
The source code will be released upon acceptance.

\end{abstract}


\section{Introduction}

\IEEEPARstart{S}{cene} flow is the dense, three-dimensional motion field that maps every point of a source observation to its corresponding location in a temporally subsequent observation referred to as the target~\cite{vedula1999}. For a vehicle navigating a dynamic environment, an accurate scene flow estimate answers two coupled questions simultaneously \emph{: What is moving} and \emph{where is it moving to} and underpins downstream tasks such as open-world object discovery~\cite{lentsch2024union}, motion forecasting~\cite{najibi2022motion}, BEV feature alignment~\cite{wang2026asyncbev} and dynamic occupancy mapping~\cite{asghar2024flow}.

The last several years have seen scene flow estimation from LiDAR converge almost entirely on deep learning. Supervised feed-forward networks regress per-point flow directly from raw points or voxelized representations~\cite{flownet3d,FastFlow3D,zhang2024deflow,kim2024flow4d}, while test-time optimization methods fit a flow field to each sweep pair using a neural prior~\cite{li2021nsfp}. These approaches deliver strong benchmark numbers, but inherit the well-known costs of supervised learning: they require large quantities of annotated LiDAR, which is expensive to produce and scarce relative to image data, they are sensitive to domain shift across sensors and environments, and optimization-based variants are far too slow for online use~\cite{vedder2024zeroflow}. 


A growing body of work puts in question the necessity of supervised learning for this scene flow.
Chodosh~\emph{et al.}~\cite{chodosh2024} re-evaluate LiDAR scene flow on realistic data and find that classical Iterative Closest Point (ICP) removes ego-motion better than any learned component, and that most of the headline gains attributed to learning actually stem from non-learned pre- and post-processing, ground removal, and piecewise-rigid refinement.
Lin~\emph{et al.}~\cite{lin2024icpflow} further demonstrated that a carefully designed ICP pipeline can achieve performance comparable to state-of-the-art learning-based methods.
In the same spirit, Khatri~\emph{et al.}~\cite{khatri2024} highlights that combining a pretrained detector with a simple tracker can match or even outperform sophisticated neural networks when the evaluation is class- and speed-aware.
Collectively, these findings suggest that effective motion estimation depends less on complex supervised models than previously believed, motivating renewed interest in unsupervised approaches.

Existing unsupervised LiDAR scene flow methods typically consist of two key components: a feed-forward neural network that regresses per-point motion and a self-supervised objective built upon the Chamfer distance between consecutive point clouds.
While this framework has underpinned the success of unsupervised LiDAR scene flow methods, its inherent drawbacks have become a bottleneck to further progress.
In particular, supervision based on the Chamfer distance is noisy, as the association between the source and target points can be erroneous when calculating the point-wise Euclidean distance solely.
SeFlow~\cite{zhang2024seflow} has observed this limitation through the deceleration of the error reduction as the training set size increases.
Given this constraint, SynFlow~\cite{zhang2026synflow} advocates for scaling up synthetic data as a solution.
By incurring the substantial cost of generating a massive synthetic dataset from simulations (34 times larger than nuScenes~\cite{caesar2020nuscenes}) and training on it, SynFlow successfully closes the gap with fully supervised learning.
While scaling up compute is widely considered the solution to many challenges in robotics, we ask: \emph{Is scaling up the only way forward for LiDAR scene flow?}

We answer with \emph{CorrelationFlow}, a novel learning-free method for LiDAR scene flow that represents a fundamental departure from existing unsupervised approaches.
Instead of automatically computing per-point flow labels, we formulate scene flow estimation as the maximization of normalized cross-correlation between two images. 
Our key insight is that while a dynamic object’s position changes over a short time interval, its point cloud appearance remains relatively constant. 
When two point clouds of the same object from neighboring time steps are projected to the Bird's-Eye View (BEV), they produce two binary images where the same foreground cluster appears shifted by an amount proportional to the object’s 3D displacement. 
We estimate this shift by identifying the placement of the first image’s foreground cluster in the second image that maximizes pixel overlap. 
This shift directly provides the object’s 3D displacement.
Under the assumption that objects only undergo rigid and translational motion over short time intervals, the 3D displacement of the object is the 3D displacement of every point on the object.
Our contributions are threefold:
\begin{enumerate}
\item \textbf{CorrelationFlow}, a fully learning-free method that formulates LiDAR scene flow estimation as a correlation maximization problem;
\item a keypoint-based variant enabling sparse scene flow computation without prior cluster knowledge;
\item a demonstration of state-of-the-art performance among unsupervised methods across diverse datasets, all while maintaining real-time runtime.
\end{enumerate}
CorrelationFlow ranked 2nd on the Argoverse 2 Scene Flow Challenge (unsupervised track) at CVPR 2026, thereby offering a new perspective on unsupervised LiDAR scene flow.

\section{Related Work}
\label{sec:related_work}

\subsection{Learning-based scene flow estimation}

Scene flow was originally introduced as the 3D counterpart of optical flow~\cite{vedula1999}. 
Early learning-based methods for LiDAR scene flow, including~\cite{flownet3d, behl2019pointflownet, guHPLFlowNet2019, wu2020pointpwc}, directly regress point-wise motion from raw point clouds using feed-forward neural architectures. 
Subsequent work improves accuracy and efficiency through more computationally efficient network designs~\cite{jund2022scalable} and by incorporating motion rigidity priors into the learning process~\cite{huang2022dynamic}. 
More recently, methods such as Flow4D~\cite{kim2024flow4d} and DeltaFlow~\cite{zhang2025deltaflow} have demonstrated that aggregating temporal information across multiple frames further improves scene flow estimation. 
Although these supervised approaches achieve state-of-the-art performance, they rely on large-scale annotated datasets for training. In contrast, our method requires no manually annotated scene flow labels.

To alleviate the reliance on annotated data, self-supervised scene flow methods have attracted increasing attention~\cite{baur2021slim, mittal2020just, vedder2024zeroflow, lin2025voteflow, zhang2026teflow}. 
Most approaches combine a feed-forward architecture for efficient inference with unsupervised objectives that eliminate the need for ground-truth supervision. 
For instance, SeFlow~\cite{zhang2024seflow} leverages cycle consistency losses~\cite{li2021nsfp} together with a PointPillars-style backbone. 
VoteFlow~\cite{lin2025voteflow} adopts the same training objectives while introducing a backbone with an embedded rigid-motion prior, motivated by the prevalence of rigid-body motion in autonomous driving.
TeFlow~\cite{zhang2026teflow} further improves accuracy by temporal ensembling across multiple frames. 
A complementary direction explores pretraining on synthetic datasets to improve generalization to real-world data~\cite{jin2022deformation, zhang2026synflow}. 
Despite eliminating the need for manual annotations, these methods still require computationally expensive training on large-scale datasets.

An alternative paradigm is test-time optimization, which removes the need for offline training altogether. Representative methods~\cite{li2021nsfp, vedder2024eulerflow, hoffmann2025floxels} optimize a scene-specific model, typically a multi-layer perceptron, directly on each input sequence by minimizing unsupervised objectives such as cycle consistency. While this avoids dataset-scale training, it incurs substantial optimization cost during inference. In contrast, our method eliminates both large-scale training and per-scene optimization.

\subsection{Learning-free scene flow estimation}
A complementary, largely learning-free line of work exploits the piecewise-rigid nature of autonomous driving scenes by first partitioning the scene into object-level clusters and then estimating a single rigid motion for each cluster~\cite{dewan2016rigid, gojcic2021, li2022rigidflow, khatri2024, chodosh2024, lin2024icpflow}. 
ICP-Flow~\cite{lin2024icpflow} further highlights the effectiveness of this geometric prior through a carefully engineered pipeline, particularly for vehicle motion estimation.

However, existing methods typically rely on density-based clustering, such as DBSCAN or HDBSCAN~\cite{campello2013hdbscan, mcinnes2017hdbscan}, which is sensitive to point density, struggles with nearby objects, and requires sensor-specific parameter tuning. These limitations become more pronounced for sparse objects and crowded scenes. 
Our work preserves the piecewise-rigid prior while replacing density-based clustering with connected components on BEV occupancy maps, providing a parameter-light and density-robust grouping strategy.

\subsection{Correlation in digital image processing}
Correlation has been widely adopted in scene flow estimation~\cite{wu2020pointpwc, wei2021pv, lin2025voteflow} for template matching across temporal observations. 
Inspired by these approaches, we use correlation to identify collective motion of pixels in the BEV domain~\cite{bolme2010mosse, henriques2015kcf}. 
Our method combines well-established components, including LiDAR BEV projection~\cite{lang2019pointpillars}, connected-component analysis~\cite{samet1988efficient, sautier2026clustering}, and distance transforms~\cite{borgefors1986distance}, without relying on learned features or descriptors.

\section{CorrelationFlow} \label{sec:correlationflow}

In this section, we first define the scene flow estimation problem. We then present the core components of the proposed method and its underlying assumptions, using the example of estimating scene flow for a single object. Finally, we generalize this formulation into a framework for estimating the scene flow of multiple objects.

\subsection{Problem statement} \label{sec:prob_stat}
We define the body frame $\ego$ of the ego vehicle by placing its origin at the ego vehicle's LiDAR sensor, with the X axis pointing along the vehicle's heading direction and the Z axis pointing opposite gravity.
Consider a 3D point $\vector{p}$ in the LiDAR point cloud, sampled from the surface of an object $\mathcal{O}$.
Let $\vector{p}^{\ego}_t \in \R^3$ denote the coordinates of $\vector{p}$ in $\ego$ at the time step $t$, and let $\vector{p}^{\ego}_{t+1} \in \R^3$ denote the coordinates of the same physical surface point at the time step $(t+1)$. 
The scene flow $\vector{f}_{t \rightarrow t+1}$ of $\vector{p}$ is defined as its displacement between the two time steps, expressed in the body frame of the ego vehicle at time $(t+1)$:

\begin{equation}
    \vector{f}_{t \rightarrow t+1} = 
    \begin{bmatrix}
        \vector{p}^{\ego}_{t+1} \\
        1
    \end{bmatrix}
    - 
    (\egotf_{t+1})^{-1} \cdot \egotf_{t} \cdot 
    \begin{bmatrix}
        \vector{p}^{\ego}_{t} \\
        1
    \end{bmatrix}
    \label{eq:def_scene_flow}
\end{equation}
where $\egotf_{t} = \begin{bmatrix}
    \rot_t & \trans_t \\ \vector{0}_{1 \times 3} & 1
\end{bmatrix} \in SE(3)$ denotes the pose of the ego vehicle with respect to the global frame $\glob$ at time step $t$, with $\rot_t \in SO(3)$ and $\trans_t \in \R^3$ being the rotational and translational components, respectively.
The operator $\cdot$ denotes matrix multiplication and $\begin{bmatrix}
    \vector{p}^{\ego}_{t} \\
    1
\end{bmatrix} \in \R^4$  
is the homogeneous coordinate of $\vector{p}^{\ego}_t$.

Eq.~\eqref{eq:def_scene_flow} implies that the ego-motion is compensated prior to the computation of scene flow.
Consequently, the scene flow $\vector{f}_{t \rightarrow t+1}$ is non-zero only if the motion of the object $\mathcal{O}$ differs from that of the ego vehicle.
We note that the scene flow defined in Eq.~\eqref{eq:def_scene_flow} is also referred to as \textit{residual flow} in the literature \cite{zhang2024seflow, zhang2024deflow, khoche2025ssf, zhang2026teflow, khoche2026dogflow}.

Let $\pc{t} = \{\vector{p}_{j,t} = (x_{j,t}, y_{j,t}, z_{j,t})\}_{j=1}^{N_t}$ denote the point cloud acquired at time step $t$, where $N_t$ is the number of points.
Given two consecutive ego-motion-compensated point clouds $\pc{t}$ and $\pc{t+1}$, our goal is to estimate the scene flow $\vector{f}_{j,t \rightarrow t+1}$ of every point $\vector{p}_{j,t} \in \pc{t}$.
We denote by $\flow{t} = \{\vector{f}_{j,t \rightarrow t+1}\}_{j=1}^{N_t}$ the scene flow vectors of all points in $\pc{t}$.

Consistent with prior works \cite{lin2024icpflow, zhang2024seflow, zhang2024deflow, khoche2025ssf, lin2025voteflow, zhang2026teflow, zhang2026synflow}, we assume that ground points are removed through preprocessing before scene flow estimation.

\subsection{Estimating the scene flow of one object}

Let $\pc{t}^{\mathcal{O}} \subset \pc{t}$ and $\pc{t+1}^{\mathcal{O}} \subset \pc{t+1}$ denote the points belonging to an object $\mathcal{O}$ at time steps $t$ and $t+1$.
Following the convention from the previous section, both point clouds are compensated for ego motion and are thus expressed in the ego vehicle's body frame at time step $t+1$.
We make the following two assumptions about object motion:
First, objects undergo rigid and predominantly planar motion, which holds for most traffic participants in autonomous driving scenarios \cite{dewan2016rigid, gojcic2021, wu2020motionnet, FastFlow3D}.
Second, during the short temporal interval between consecutive point clouds, object motion can be approximated as pure translation.
Thus, all points in $\pc{t}^{\mathcal{O}}$ share the same scene flow $\vector{f}_{t \rightarrow t+1} = [f_x, f_y, f_z]^\top \in \R^3$.

To estimate the scene flow $\vector{f}_{t \rightarrow t+1}$ of $\pc{t}^{\mathcal{O}}$, we orthogonally project the point clouds onto 2D planes (e.g., the horizontal plane), referred to as *views*.
This approach leverages the equivariance of planar rigid motion under such projections.
Moreover, the resulting 2D representation allows us to analyze point clouds using image processing tools.
Since $\vector{f}_{t \rightarrow t+1}$ has three components but each view only provides a 2D estimate, at least two views are required.
We estimate $\vector{f}$ using two mutually orthogonal views: the BEV and the sectional view.
The BEV is the projection onto the plane spanned by the X and Y axes of the ego vehicle's body frame at time step $t+1$.
The sectional view is the projection onto the vertical plane aligned with the object $\mathcal{O}$'s heading direction.
The BEV provides the horizontal components $[f_x, f_y]^\top$, while the sectional view additionally recovers the vertical component $f_z$.

\begin{figure}[b!]
    \centering
    \vspace{-5mm}
    \begin{subfigure}[t]{0.49\linewidth}
        \includegraphics[trim=0cm 2cm 0cm 0cm, clip,width=\linewidth]{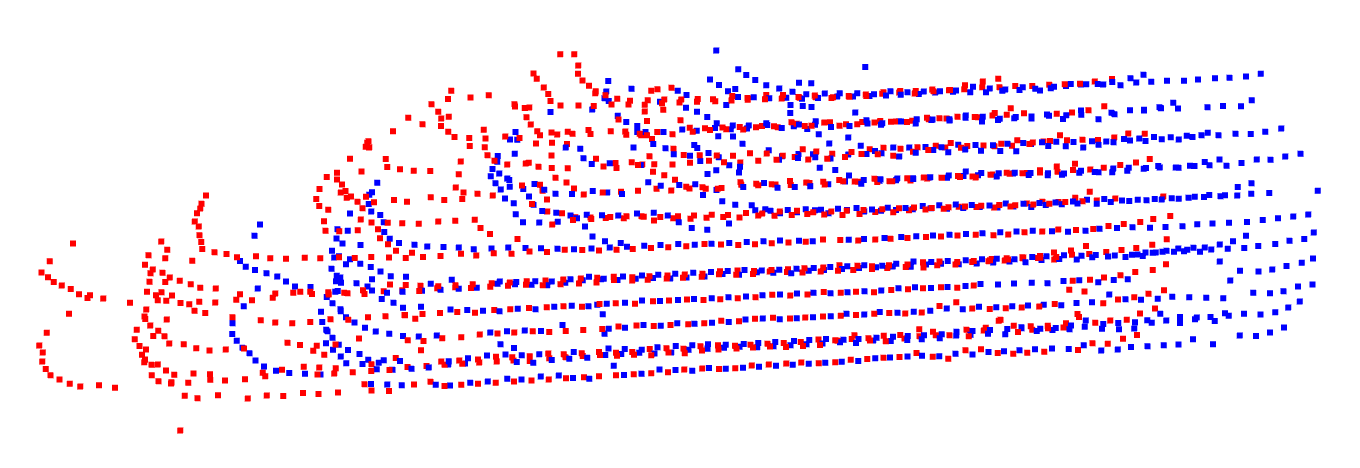}
        \caption{\scriptsize $\pc{t}^{\mathcal{O}}$ (blue) and $\pc{t+1}^{\mathcal{O}}$ (red).}
        \label{fig:sfl_1obj_pointclouds}
    \end{subfigure}
    \hfill
    \begin{subfigure}[t]{0.49\linewidth}
        \includegraphics[trim=1.5cm 4cm 1.5cm 3cm, clip, width=\linewidth]{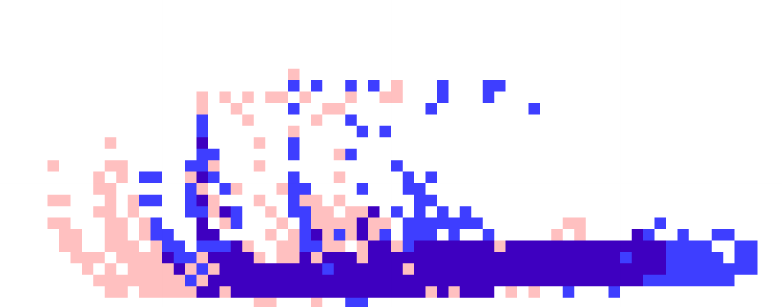}
        \caption{\scriptsize $\img{t}^{\mathcal{O}}$ (blue) and $\img{t+1}^{\mathcal{O}}$ (red).}
        \label{fig:sfl_1obj_bev}
    \end{subfigure}

    \begin{subfigure}[t]{0.49\linewidth}
        \includegraphics[trim=0cm 2cm 0cm 1cm, clip, width=\linewidth]{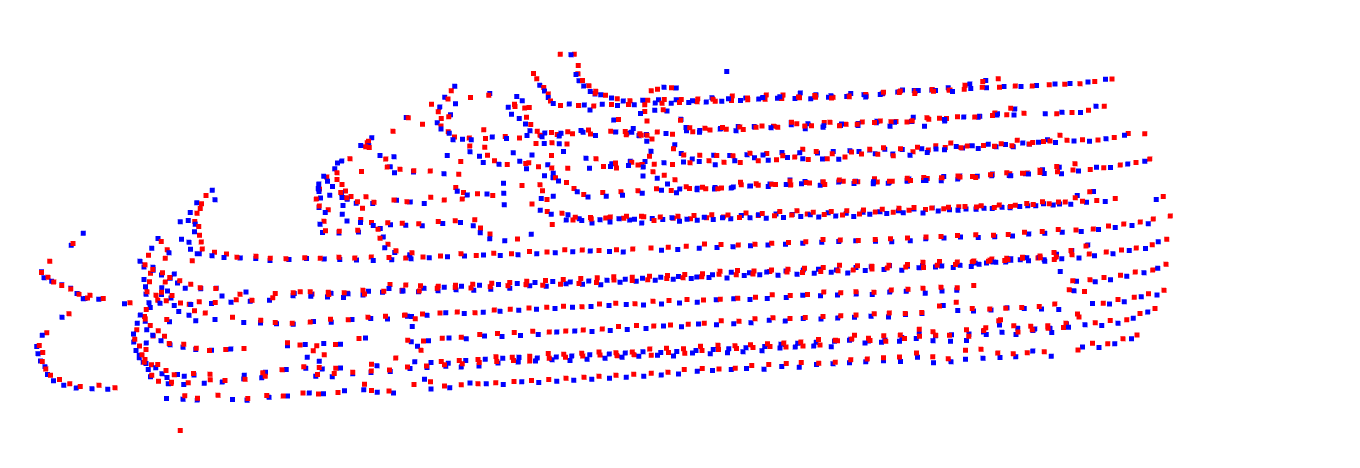}
        \caption{\scriptsize $\pc{t}^{\mathcal{O}}$ translated by $\vector{f}_{t \rightarrow t+1}$ and $\pc{t+1}^{\mathcal{O}}$.}
        \label{fig:sfl_1obj_pointclouds_with_flow}
    \end{subfigure}
    \hfill
    \begin{subfigure}[t]{0.49\linewidth}
        \includegraphics[trim=1.5cm 4cm 1.5cm 3cm, clip, width=\linewidth]{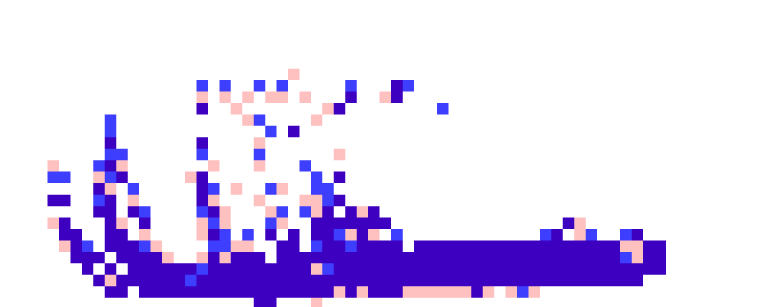}
        \caption{\scriptsize $\img{t}^{\mathcal{O}}$ translated by $\vector{\Delta}$ and $\img{t+1}^{\mathcal{O}}$.}
        \label{fig:sfl_1obj_bev_warped}
    \end{subfigure}
    \caption{\small Point cloud of a dynamic car at time step $t$ (blue) and $(t+1)$ (red), expressed in the body frame of the ego vehicle at time step $(t+1)$, and its projection onto the BEV (a, b). Translating the point cloud by the scene flow $\vector{f}_{t \rightarrow t+1}$, and correspondingly the image by $\vector{\Delta}$, aligns the two time steps (c, d).}
    \vspace{-4mm}
\end{figure}

We start with the BEV to estimate the horizontal components $[f_x, f_y]^\top$ of $\vector{f}_{t \rightarrow t+1}$.
The orthogonal projection of the 3D point $\vector{p} = [p_x, p_y, p_z]^\top$ onto the BEV is given by
\begin{equation}
    \vector{u} = \left\lfloor \frac{(p_x, p_y) - (p_{x}^{-}, p_{y}^{-})}{s} \right\rfloor,
    \label{eq:3d_to_bev}
\end{equation}
where $s$ is the quantization factor in meters, $\lfloor \cdot \rfloor$ is the element-wise floor operator, and the vector $(p_{x}^{-}, p_{y}^{-})$ denotes the lower bound of the region of interest in the horizontal plane.
This factor $s$ discretizes the BEV plane into a 2D grid of $H \times W$ cells, with
\begin{equation}
    H = \left\lfloor \frac{p_{y}^{+} - p_{y}^{-}}{s} \right\rfloor + 1, \quad
    W = \left\lfloor \frac{p_{x}^{+} - p_{x}^{-}}{s} \right\rfloor + 1.
    \label{eq:bev_dims}
\end{equation}
Each cell (referred to as a pixel) has size $s \times s$.
We assign the value $1$ to pixels occupied by at least one projected point and $0$ to free pixels to convert the projection of $\pc{t}^{\mathcal{O}}$ into a binary image $\img{t}^{\mathcal{O}} \in \{0, 1\}^{H \times W}$.
Since both point clouds use the same region of interest, the images $\img{t}^{\mathcal{O}}$ and $\img{t+1}^{\mathcal{O}}$ have identical dimensions.

Following the projection of individual points in Eq.~\eqref{eq:3d_to_bev}, the scene flow $\vector{f}_{t \rightarrow t+1}$ projects onto the BEV as $\vector{\Delta} = [\Delta_x, \Delta_y]^\top \in \mathbb{Z}^2$, given by
\begin{equation}
    \vector{\Delta} = \left\lfloor [f_x, f_y]^\top / s \right\rfloor.
    \label{eq:vectorXY_3d_to_bev}
\end{equation}
Since $\vector{f}_{t \rightarrow t+1}$ translates every point $\vector{p}_{j,t} \in \pc{t}^{\mathcal{O}}$ to its correspondence in $\pc{t+1}^{\mathcal{O}}$ (Fig.~\ref{fig:sfl_1obj_pointclouds_with_flow}), translating $\img{t}^{\mathcal{O}}$ by $\vector{\Delta}$ approximately maximizes the overlap between its foreground and that of $\img{t+1}^{\mathcal{O}}$ (Fig.~\ref{fig:sfl_1obj_bev_warped}).
We assume that the inter-frame rotation of the object is negligible, which holds at typical LiDAR frame rates: even at a high yaw rate of $30^\circ/\mathrm{s}$, an object rotates by merely $3^\circ$ between two frames captured at 10~Hz.
Moreover, the alignment is only approximate due to the non-additivity of the floor operator, which causes a quantization error of at most one pixel per direction.
Consequently, estimating the horizontal components of $\vector{f}_{t \rightarrow t+1}$ reduces to finding the 2D translation $\vector{\Delta}$ that maximizes the overlap between the foregrounds of the two images.

We use normalized cross-correlation (NCC) \cite{lewis1995ncc} to measure the overlap in the foreground between the two binary images, with $\img{t}^{\mathcal{O}}$ serving as a template that slides over $\img{t+1}^{\mathcal{O}}$.
Let $\vector{u} = [u_x, u_y]^\top \in \mathbb{Z}^2$ denote the coordinates of a pixel in $\img{t+1}^{\mathcal{O}}$.
The NCC score $R(u_x, u_y)$ is computed over the $H \times W$ neighborhood of $\vector{u}$ as
\begin{equation}
\begin{split}
    &R(u_x, u_y) = \\ & \frac{
        \sum\limits_{r=0}^{H-1}\sum\limits_{c=0}^{W-1} 
            \img{t}^{\mathcal{O}}(r, c) \, \img{t+1}^{\mathcal{O}}(\widetilde{u}_y + r, \widetilde{u}_x + c)
    }{
        \sqrt{
            \sum\limits_{r=0}^{H-1}\sum\limits_{c=0}^{W-1} \img{t}^{\mathcal{O}}(r, c)^2
        }
        \sqrt{
            \sum\limits_{r=0}^{H-1}\sum\limits_{c=0}^{W-1} \img{t+1}^{\mathcal{O}}(\widetilde{u}_y + r, \widetilde{u}_x + c)^2
        }
    },
\end{split}
\label{eq:norm_correlation}
\end{equation}
where $(\widetilde{u}_x, \widetilde{u}_y) = (u_x - \lfloor W/2 \rfloor, u_y - \lfloor H/2 \rfloor)$ denotes the top-left corner of the neighborhood.
Pixels outside the image border are set to $0$, equivalent to zero-padding $\img{t+1}^{\mathcal{O}}$ on four sides.

\begin{figure}[b]
    \centering
    \includegraphics[trim=3cm 0cm 0.5cm 3cm, clip,width=0.95\linewidth]{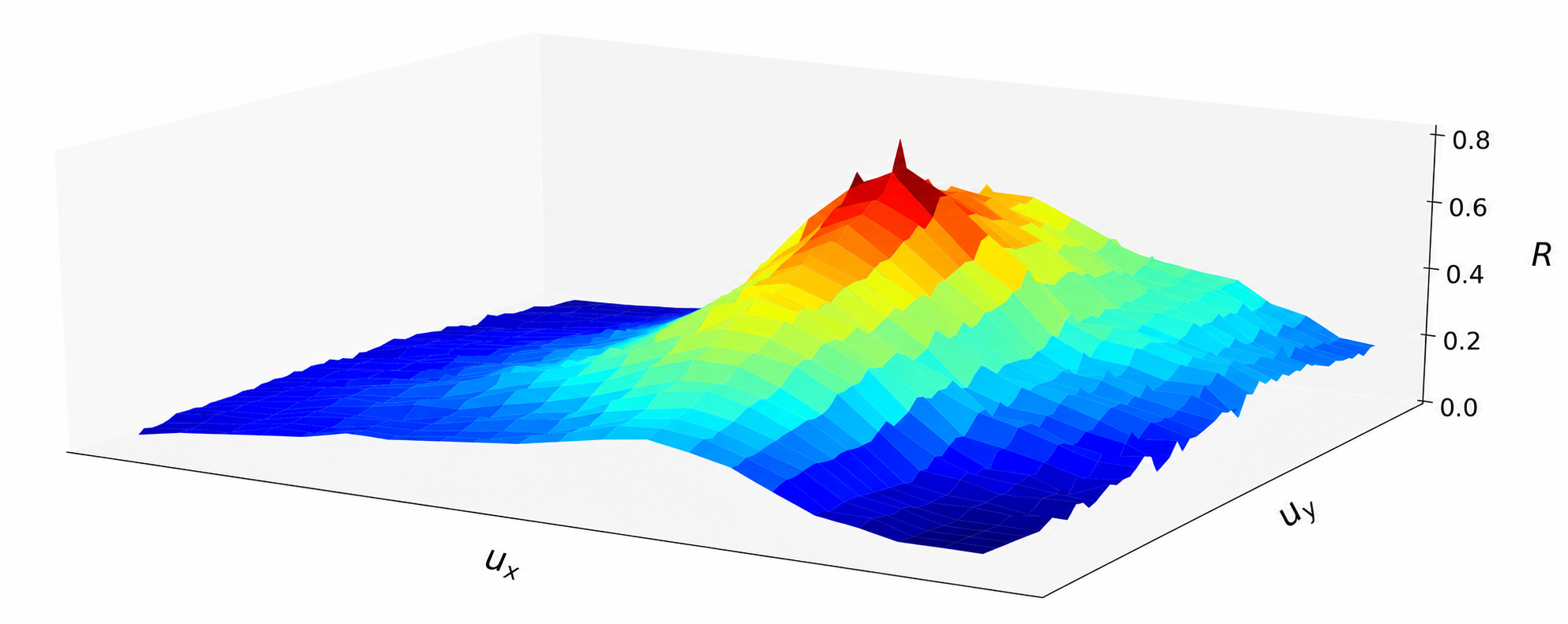}
    \caption{\small NCC map of $\img{t+1}^{\mathcal{O}}$ with respect to the template $\img{t}^{\mathcal{O}}$, rendered as a surface over the pixel coordinates $(u_x, u_y)$. The map exhibits a single distinct peak at $\vector{u}^*$, indicating the placement of the template that maximizes the foreground overlap. The BEV projection of the scene flow follows from the offset of this peak via Eq.~\eqref{eq:displacement_in_bev}.}
    \label{fig:sfl_1obj_corremap}
    \vspace{-3mm}
\end{figure}

An advantage of the NCC is that it takes values in $[0, 1]$, where $0$ indicates no overlap and $1$ indicates maximum overlap.
Fig.~\ref{fig:sfl_1obj_corremap} shows the NCC map, i.e., the $R$ value at each pixel of $\img{t+1}^{\mathcal{O}}$.
The map exhibits a distinct peak at $\vector{u}^* = \argmax_{(u_x, u_y)} R(u_x, u_y)$, which indicates the optimal placement of $\img{t}^{\mathcal{O}}$'s center in $\img{t+1}^{\mathcal{O}}$ to maximize foreground overlap.
The BEV projection $\vector{\Delta}$ of the scene flow $\vector{f}_{t \rightarrow t+1}$ is therefore the displacement of this center, i.e., the difference between $\vector{u}^*$ and the center of $\img{t}^{\mathcal{O}}$:
\begin{equation}
    \vector{\Delta} = \vector{u}^* - \left[ \lfloor W/2 \rfloor, \lfloor H/2 \rfloor \right]^\top.
    \label{eq:displacement_in_bev}
\end{equation}
Following Eq.~\eqref{eq:vectorXY_3d_to_bev}, the horizontal components of $\vector{f}_{t \rightarrow t+1}$ are recovered as $[f_x, f_y]^\top \approx s \cdot \vector{\Delta}$.
The single-view flow estimation is summarized in Alg.~\ref{alg:scene_flow_1view}.

To obtain the vertical component $f_z$, we construct the sectional view as follows.
First, we compute the heading direction of the object in the BEV from the estimated horizontal components as $\theta = \arctantwo(f_y, f_x)$.
Next, we compensate $\pc{t}^{\mathcal{O}}$ for the object's horizontal motion by translating it by $[f_x, f_y, 0]^\top$.
Both point clouds are then rotated about the Z axis by $-\theta$ to align the object's heading with the X axis.
Finally, we apply Alg.~\ref{alg:scene_flow_1view} with the X--Z plane as the chosen view to obtain $f_z$.
The complete single-object scene flow estimation procedure is summarized in Alg.~\ref{alg:scene_flow_1obj}.

\begin{algorithm}[htbp]
\small
\SetAlgoLined
\KwData{Point clouds $\pc{t}^{\mathcal{O}}$ and $\pc{t+1}^{\mathcal{O}}$ of one object at two consecutive time steps and the chosen view}
\KwResult{Components of scene flow in the chosen view}
\caption{\small Estimating the scene flow of one object in one view}
\label{alg:scene_flow_1view}
\Function{scene\_flow\_in\_1view($\pc{t}^{\mathcal{O}}, \pc{t+1}^{\mathcal{O}}, \textrm{view}$)}
{
    \vspace{0.2cm}
    
    $\img{t}^{\mathcal{O}} \leftarrow$ projection of $\pc{t}^{\mathcal{O}}$ onto $\textrm{view}$ using Eq.~\eqref{eq:3d_to_bev}

    $\img{t+1}^{\mathcal{O}} \leftarrow$ projection of $\pc{t+1}^{\mathcal{O}}$ onto $\textrm{view}$ using Eq.~\eqref{eq:3d_to_bev}

    \vspace{0.2cm}
    
    \tcc{
        \footnotesize calculate the NCC with $\img{t}^{\mathcal{O}}$ for each pixel of $\img{t+1}^{\mathcal{O}}$ 
    }

    $\mathbf{R} \leftarrow $ a matrix of size $H \times W$ filled with $0$s with $H, W$ being the height and width of $\img{t+1}^{\mathcal{O}}$

    \For{$u_y \in [0, H[$}{
        \For{$u_x \in [0, W[$}{
            $\mathbf{R}[u_y, u_x] \leftarrow R(u_x, u_y)$ with $R(u_x, u_y)$ defined in Eq.~\eqref{eq:norm_correlation}

            \tcp{$\mathbf{R}[u_x, u_y]$ \footnotesize denote the cell at row $u_y$, column $u_x$ of the matrix $\mathbf{R}$}
        }
    }

    \tcc{
        \footnotesize compute the components of scene flow in the chosen view
    }

    $[u_x^*, u_y^*] \longleftarrow \mathop{\argmax}\limits_{(u_x, u_y) \in [0, W[ \times [0, H[} \mathbf{R}$
    
    \vspace{0.2cm}
    
    $\vector{\Delta} = [u_x^*, u_y^*] - \left[\lfloor \frac{W}{2} \rfloor, \lfloor \frac{H}{2} \rfloor \right]$

    \vspace{0.2cm}
    
    \Return $\vector{\Delta} \times s$ with $s$ being the quantization factor
}
\end{algorithm}

\begin{algorithm}[b]
\small
\SetAlgoLined
\KwData{Point clouds $\pc{t}^{\mathcal{O}}$ and $\pc{t+1}^{\mathcal{O}}$ of one object at two consecutive time steps and the chosen view}
\KwResult{scene flow $\vector{f}_{t \rightarrow t+1} = [f_x, f_y, f_z]^{\top}$}
\vspace{0.15cm}
\caption{\small Estimating the 3D scene flow of one object}
\label{alg:scene_flow_1obj}
\Function{scene\_flow\_1obj($\pc{t}^{\mathcal{O}}, \pc{t+1}^{\mathcal{O}}$)}
{
    \vspace{0.2cm}
    
    $[f_x, f_y]^{\top} \leftarrow $ scene\_flow\_in\_1view($\pc{t}^{\mathcal{O}}, \pc{t+1}^{\mathcal{O}}$, view=\textit{BEV})

    \vspace{0.15cm}

    Translate $\pc{t}^{\mathcal{O}}$ by $[f_x, f_y, 0]^{\top}$

    \vspace{0.15cm}
    
    $\theta \leftarrow -\arctan(f_y, f_x)$

    Rotate the two point clouds $\pc{t}^{\mathcal{O}}$ and $\pc{t+1}^{\mathcal{O}}$ around the Z axis of the ego vehicle's frame by $\theta$

    \vspace{0.15cm}

    $[f_x', f_z]^{\top} \leftarrow $ scene\_flow\_in\_1view($\pc{t}^{\mathcal{O}}, \pc{t+1}^{\mathcal{O}}$, view=\textit{X-Z})

    $f_x \leftarrow f_x + f_x' \cdot \cos(-\theta)$

    $f_y \leftarrow f_y + f_x' \cdot \sin(-\theta)$

    \vspace{0.2cm}
    \Return $[f_x, f_y, f_z]^{\top}$
}
\end{algorithm}

\subsection{Estimating the scene flow of multiple objects}
The previous section relies on the rigidity assumption to formulate the estimation of scene flow for a single object, given its point clouds $\pc{t}^{\mathcal{O}}$ and $\pc{t+1}^{\mathcal{O}}$, as the maximization of normalized cross-correlation between their projected images. 
This formulation assumes rigidity and pure translation over short time intervals. 
To extend this to the full scene point clouds $\pc{t}$ and $\pc{t+1}$, we partition them into spatio-temporal clusters $\{\cluster{i}\}_{i=1}^{M}$, where each cluster $\cluster{i} = \pc{t}^{i} \cup \pc{t+1}^{i}$ groups points originating from the same physical object at time steps $t$ and $t+1$. 
By construction, points within a cluster satisfy the rigidity assumption, so we can apply Alg.~\ref{alg:scene_flow_1obj} to each cluster individually to obtain its scene flow. 
The construction of these spatio-temporal clusters is detailed in the next section, and the full CorrelationFlow pipeline is summarized in Alg.~\ref{alg:general_correlation_flow}.

\begin{algorithm}[htbp]
\small
\SetAlgoLined
\KwData{A point cloud sequence $\mathcal{S} = \{\pc{t-h}, \dots, \pc{t}, \pc{t+1}\}$}
\KwResult{Scene flow $\flow{t}$}
\vspace{0.1cm}
Partition point clouds $\pc{t}$ and $\pc{t+1}$ into a set of spatio-temporal clusters $\{\cluster{i} = \pc{t}^{i} \cup \pc{t+1}^{i}\}$ where $\pc{t}^{i}$ and $\pc{t+1}^{i}$ contain points of the cluster $i$ at time step $t$ and $t+1$, respectively

$\flow{t} \leftarrow \emptyset$

\For{each cluster $\cluster{i}$}{
    \tcc{estimate per cluster scene flow}
    $\vector{f}_{t \rightarrow t+1}^i \leftarrow$ scene\_flow\_1obj$(\pc{t}^{i}, \pc{t+1}^{i})$

    \vspace{0.2cm}
    \tcc{assign $\vector{f}_{t \rightarrow t+1}^i$ to points of $\pc{t}^{i}$}
    
    $\flow{t} \leftarrow \flow{t} \cup \left\{  \vector{f}_{t \rightarrow t+1}^i \right\}_{j=1}^{|\pc{t}^{i}|}$
}

\vspace{0.1cm}
\Return{$\flow{t}$}
\caption{\small Scene flow by correlation maximization}
\label{alg:general_correlation_flow}
\end{algorithm}

\vspace{-5mm}

\section{Components of CorrelationFlow} \label{sec:correaltion_flow_components}
This section presents the components that refine the core formulation of Sec.~\ref{sec:correlationflow}.
These components address the challenge of detecting spatio-temporal clusters of objects, which can have disjoint parts due to occlusion, sparsity in point clouds, or high-speed motion.
They also leverage multiple past point clouds to improve scene flow estimation.
\vspace{-2mm}
\subsection{Spatio-temporal clustering using connected components} \label{sec:clustering}

\begin{figure}[t]
    \centering
    \includegraphics[width=0.8\linewidth]{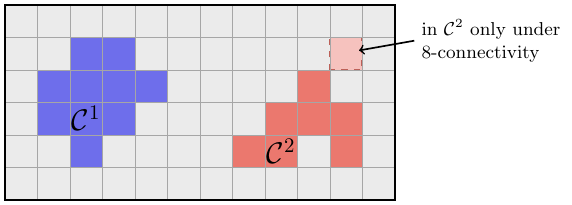}
    \caption{\small Two connected components $\cc^{1}$ and $\cc^{2}$ of a binary image. Foreground pixels are grouped into a component if they are directly or transitively connected. The dashed pixel touches $\cc^{2}$ only diagonally and is included in the component only under 4-connectivity.}
    \label{fig:example_connected_comp}
       \vspace{-3mm}

\end{figure}

We propose a spatio-temporal clustering approach that processes an entire point cloud sequence in a single pass with minimal hyperparameters.
Our method builds upon the notion of connectivity in binary images: pixel $\vector{v}$ is connected to pixel $\vector{u}$ if $\vector{v}$ lies in the immediate neighborhood of $\vector{u}$, which comprises the four orthogonally adjacent pixels and, optionally, the four diagonally adjacent ones.
Our key observations are twofold.
First, in the BEV at a given time step, the parts of an object form a connected region, as illustrated in Fig.~\ref{fig:sfl_1obj_bev}.
Second, after ego-motion compensation, the footprints of the same object at consecutive time steps overlap or adjoin and are thus also connected.
Based on these observations, we identify spatio-temporal clusters in a point cloud sequence $\mathcal{S} = \{\pc{t-h}, \dots, \pc{t}, \pc{t+1}\}$ in four steps:
\begin{enumerate}
    \item compensating every point cloud in $\mathcal{S}$ for ego-motion,
    \item aggregating the compensated point clouds into a single point cloud $\pc{\mathrm{agg}}$,
    \item projecting $\pc{\mathrm{agg}}$ onto the BEV using Eq.~\eqref{eq:3d_to_bev} to obtain the binary image $\img{\mathrm{agg}}$,
    \item extracting connected components from $\img{\mathrm{agg}}$ using the algorithm of \cite{bolelli2022connected}.
\end{enumerate}
Each connected component $\cc^{i}$ is a binary image of the same size as $\img{\mathrm{agg}}$, with foreground pixels that are directly or transitively connected to each other but not to the remaining foreground of $\img{\mathrm{agg}}$.
Fig.~\ref{fig:example_connected_comp} illustrates two connected components identified in a binary image.

\begin{figure}[b]
    \centering
    \vspace{-6mm}
    \includegraphics[width=0.9\linewidth]{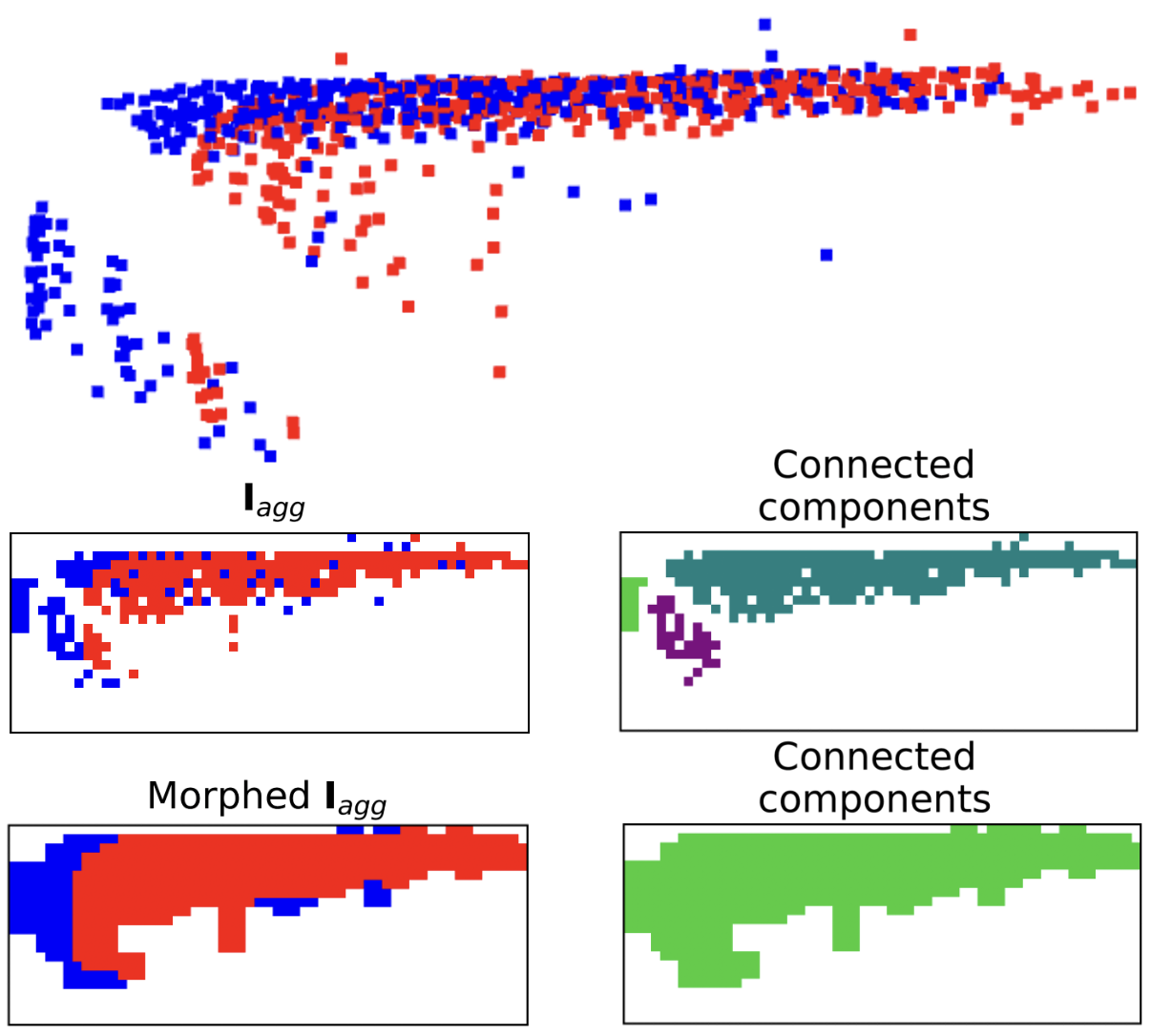}
    \caption{\small Effect of dilation on the clustering of an object with disjoint parts. Top: point cloud of a vehicle whose BEV footprint is split by occlusion and sparsity, at time step $t$ (blue) and $(t+1)$ (red). Middle: the aggregated image $\img{\mathrm{agg}}$ and its connected components; the disjoint parts yield three separate components. Bottom: the dilated $\img{\mathrm{agg}}$ and its connected components; the dilation bridges the gaps, merging the object into a single component.}

    \label{fig:clustering_obj_with_disjoint_parts}

\end{figure}

Occlusion and sparsity, inherent to LiDAR point clouds, can split large or distant objects into disjoint parts in the BEV.
To bridge such gaps, we apply morphological dilation with a $3 \times 3$ structuring element of ones, $\mathbf{1}_{3 \times 3}$, to $\img{\mathrm{agg}}$ before extracting connected components.
Fig.~\ref{fig:clustering_obj_with_disjoint_parts} compares the clustering of an object with disjoint parts between the original image $\img{\mathrm{agg}}$ and its dilated version, the dilation inserts foreground pixels between the vehicle's parts, merging them into a single connected component.

The dilation can also falsely join nearby objects, producing abnormally large connected components.
To address this, we add a validation step: if the longer side of a connected component's bounding box exceeds a threshold $d^{\max}$ (reflecting the maximum expected object size), we erode the component with the same structuring element $\mathbf{1}_{3 \times 3}$ to remove most pixels introduced by the dilation.
We then re-extract connected components from the eroded component image $\img{\mathrm{agg}}^{\cc}$.
The complete spatio-temporal clustering procedure is summarized in Alg.~\ref{alg:clustering}.

\begin{algorithm}[htbp]
\small
\SetAlgoLined
\KwData{A point cloud sequence $\mathcal{S} = \{\pc{t-h}, \dots, \pc{t}, \pc{t+1}\}$}
\KwResult{
    The set of spatio-temporal clusters \\
    $\left\{\cluster{i} =  \pc{t-h}^i \cup \dots \cup \pc{t}^i \cup \pc{t+1}^i  \right \}$
    such that
    $\mathcal{S} = \bigcup_i \cluster{i}$
}

\vspace{0.25cm}
\tcc{\footnotesize ego motion compensation \& aggregation}
$\pc{agg} \leftarrow \emptyset $

\For{$\pc{k}$ in $\mathcal{S}$}{
    Transform $\pc{k}$ to the frame of $\pc{t+1}$ 
    
    $\pc{agg} \leftarrow \pc{agg} ~ \cup ~ \pc{k}$
}

\vspace{0.25cm}
$\img{agg} \leftarrow $ projection of $\pc{agg}$ to the BEV plane

Dilate $\img{agg}$ using $\mathbf{1}_{3\times3}$ kernel

$\left\{ \mathcal{C}^i \right\} \leftarrow $ connected components in $\img{agg}$
\tcc{$\mathcal{C}^i$ \footnotesize is a binary image of the same size as $\img{agg}$}

\vspace{0.25cm}
\tcc{\footnotesize validation of connected components}
$\mathcal{D} \leftarrow \left\{ \mathcal{C}^i \right\}$

$\mathcal{D}^{\text{new}} \leftarrow \emptyset$

\For{\text{each connected component} $\mathcal{C}^i$ in $\mathcal{D}$}{
    
    $(h, w) \leftarrow $ the size of the bounding box of foreground pixels of $\mathcal{C}^i$
    
    \If{$h > d^{\text{max}}$ or $w > d^{\text{max}}$}
    {
        Erode $\cc^i$ using $\mathbf{1}_{3\times3}$ kernel

        Detect connected components in eroded $\cc^i$

        Add newly detected connected components to $\mathcal{D}^{\text{new}}$


        Remove $\mathcal{C}^i$ from $\mathcal{D}$
    }
}

$\mathcal{D} \leftarrow \mathcal{D} ~ \cup ~ \mathcal{D}^{\text{new}}$

\vspace{0.25cm}
\tcc{\footnotesize assign points to clusters}
\For{\text{each connected component} $\mathcal{C}^i$ in $\mathcal{D}$}{
    $\cluster{i} \leftarrow \emptyset$

    \For{\text{each 3D point} $\vector{p} \in \mathcal{S}$}{
        $\cluster{i} \leftarrow \cluster{i} ~ \cup ~ \vector{p}$ if the projection of $\vector{p}$ to the BEV (Eq.~\eqref{eq:3d_to_bev}) is a foreground pixel of $\mathcal{C}^i$
    }
}

\vspace{0.1cm}
\Return{
    $\left\{\cluster{i} =  \pc{t-h}^i \cup \dots \cup \pc{t}^i \cup \pc{t+1}^i \right \}$
}

\caption{\small Spatio-temporal clustering using connected components}
\label{alg:clustering}
\end{algorithm}

\subsection{Multi-scale strategy}
\begin{figure}[t]
    \centering
    \includegraphics[width=0.99\linewidth]{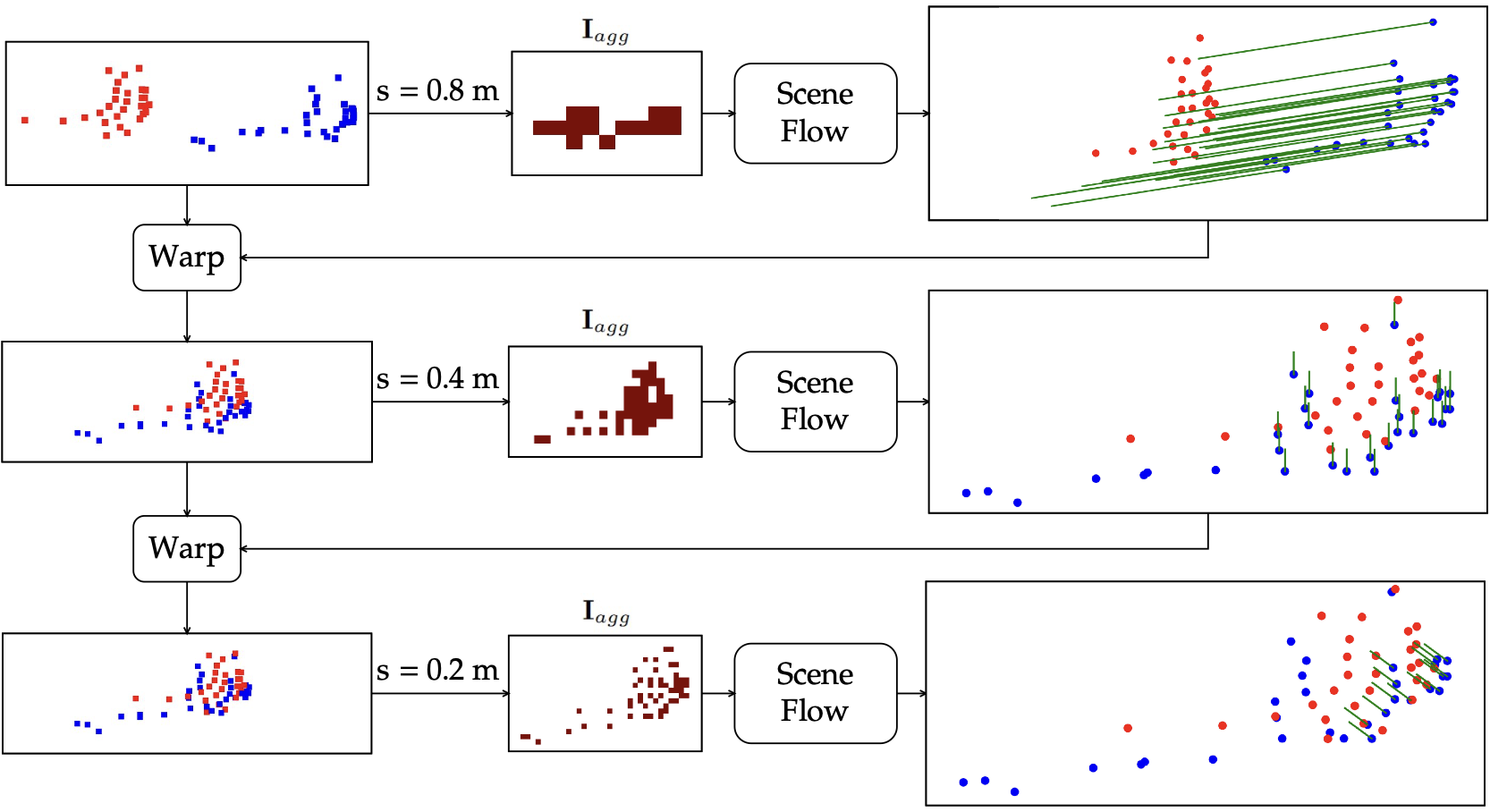}
    \caption{\small Coarse-to-fine estimation for fast-moving objects: the flow estimated at each pixel size $s$ warps $\pc{t}$ (blue) toward $\pc{t+1}$ (red) before the next, finer level.}
    \label{fig:multicale_strategy}
\end{figure}

Fast-moving objects can travel farther than their own footprint between time steps, potentially leaving their BEV instances disconnected and thus split by the clustering.
To handle such objects, we employ a coarse-to-fine strategy, illustrated in Fig.~\ref{fig:multicale_strategy}.
We first apply Alg.~\ref{alg:general_correlation_flow} at a coarse resolution, i.e., with a large pixel size $s$, where the footprints of the same object are more likely to remain connected.
We then translate $\pc{t}$ by the estimated flow and repeat the process at progressively finer resolutions until reaching the finest resolution.

\vspace{-5mm}
\subsection{Leveraging past point clouds} \label{sec:leverage_past_pcs}
Each cluster $\cluster{i}$ identified by Alg.~\ref{alg:clustering} contains the 3D points of an object across time steps $\{t-h, \dots, t, t+1\}$.
For any time steps $m, n$ with $t-h \leq m < n \leq t+1$, Alg.~\ref{alg:scene_flow_1obj} estimates the displacement $\vector{f}_{m \rightarrow n}$ of the cluster.
Assuming constant velocity, dividing $\vector{f}_{m \rightarrow n}$ by $(n - m)$ yields an estimate of $\vector{f}_{t \rightarrow t+1}$.
Thus, multiple pairs $(m, n)$ provide multiple estimates, which we pool to reduce estimation error.
We skip pairs where the cluster is empty at either time step.
In our implementation, we compute $\vector{f}_{m \rightarrow t+1}$ for every $m \in \{t-h, \dots, t\}$, normalize each vector by $(t+1-m)$, and pool the normalized estimates by taking the median of their magnitudes and orientations separately.

\vspace{-5mm}

\section{CorrelationFlow-Keypoints}
A key component of CorrelationFlow (Sec.~\ref{sec:correlationflow}) is the identification of spatio-temporal clusters in the input point clouds. Together with the rigid-motion assumption, these clusters provide a strong constraint: all points in a cluster share the same scene flow. Several works exploit this same constraint to estimate or refine scene flow \cite{dewan2016rigid, li2022rigidflow, lin2024icpflow, lin2025voteflow}.

However, clustering itself is challenging and error-prone. 
As discussed in Sec.~\ref{sec:related_work}, its quality degrades with sparse, distant, or nearby objects, and errors in cluster assignment propagate directly to the estimated flow. 
This leads us to question whether explicit clustering is necessary for scene flow estimation. 
We therefore develop a variant, CorrelationFlow-KeyPoints, that removes the clustering step from Alg.~\ref{alg:general_correlation_flow} and instead estimates the motion of individual points. 
This variant also operates on a single sweep pair.

Formally, given the BEV images $\img{t}$ and $\img{t+1}$ of point clouds $\pc{t}$ and $\pc{t+1}$, we seek the displacement of each occupied pixel in $\img{t}$. To this end, we follow the classical sparse-matching pipeline for optical flow, which consists of four steps:
\begin{enumerate}
    \item detecting key points in $\img{t}$ and $\img{t+1}$,
    \item computing a descriptor for each key point,
    \item matching the key points of $\img{t}$ to those of $\img{t+1}$ via their descriptors, and
    \item taking the coordinate difference between each key point of $\img{t}$ and its correspondence in $\img{t+1}$ as the displacement.
\end{enumerate}
The remainder of this section details our implementation of these steps.
\vspace{-3mm}
\subsection{Keypoints}

\begin{figure}[t!]
    \centering
    \includegraphics[width=0.95\linewidth]{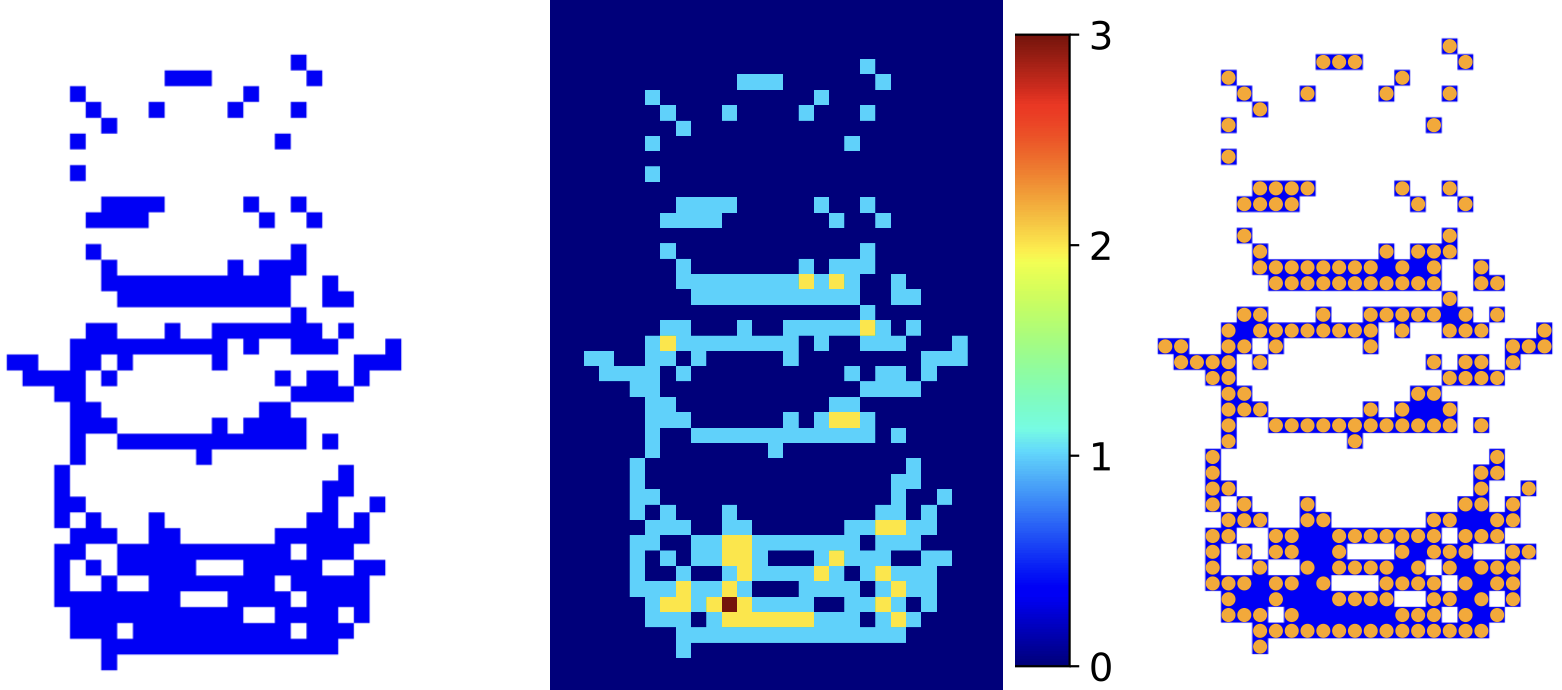}
    \caption{\small Identification of key points in the BEV image $\img{t}$ of a car. Left: the binary image $\img{t}$. Middle: the distance transform of $\img{t}$, i.e., the distance of each foreground pixel to the nearest background pixel. Right: key points (orange), selected as the boundary pixels of the foreground.}
    \label{fig:id_key_pt}
\end{figure}

Key points are defined as the pixels on the boundary of an object's BEV footprint.
As illustrated in Fig.~\ref{fig:id_key_pt}, their identification proceeds in three steps.
First, we project $\pc{t}$ onto the BEV using Eq.~\eqref{eq:3d_to_bev} to obtain the binary image $\img{t}$, with foreground pixels (value $1$) containing at least one projected point and background pixels (value $0$) being free.
To identify boundary pixels, we measure the distance from each foreground pixel to the nearest background pixel using the distance transform of $\img{t}$.
Finally, we select foreground pixels with distance $1$ to the nearest background as key points.
The density of key points adapts to the scene structure: for large, densely sampled objects, only boundary pixels are retained, while for sparse, distant objects with thin footprints, most foreground pixels qualify as key points.
This behavior is desirable, as it concentrates matching on geometrically informative pixels while naturally preserving coverage of sparsely observed objects.
\vspace{-1mm}
\subsection{Descriptor}

\begin{figure}[b!]
    \centering
    \includegraphics[width=0.95\linewidth]{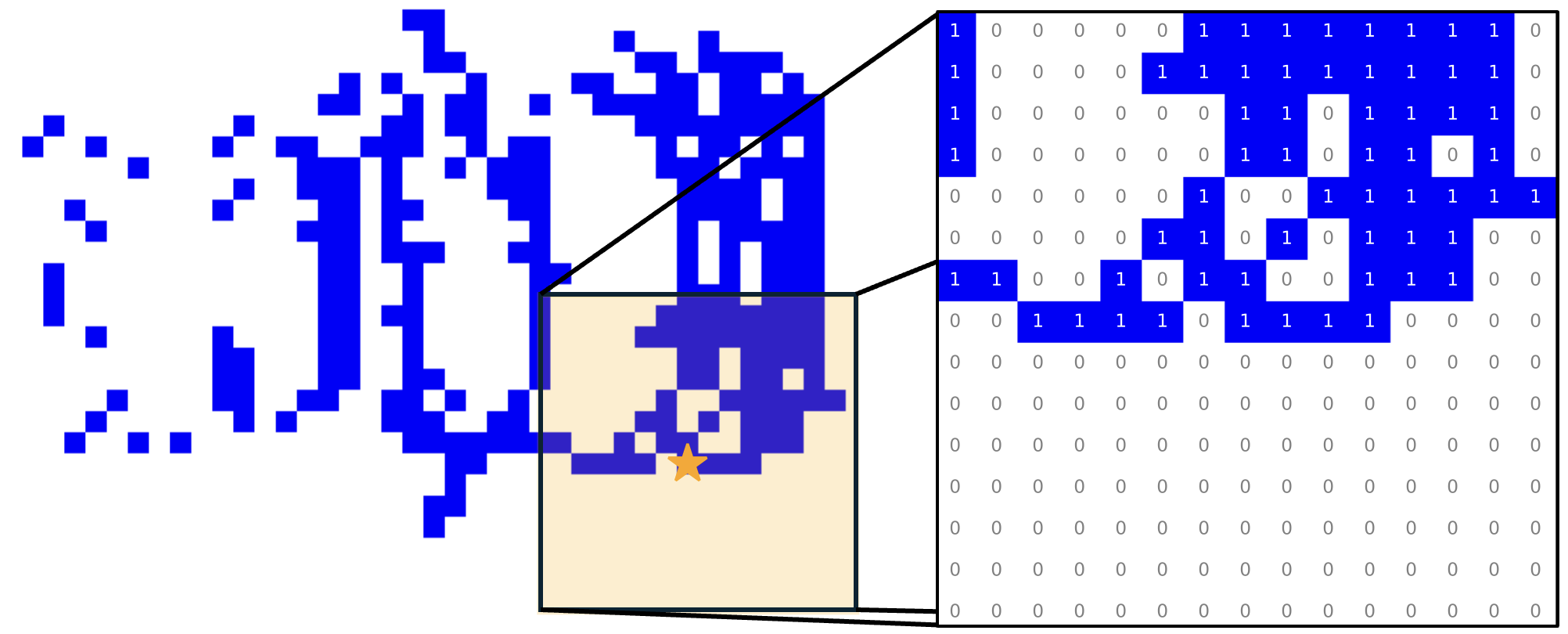}
    \caption{\small Descriptor of a key point in the BEV image $\img{t}$. The descriptor of the key point (orange star) is the $L \times L$ patch of $\img{t}$ centered at the key point (orange square), shown enlarged on the right with its binary values.}
    \label{fig:key_pt_descriptor}
\end{figure}

Given a key point $\vector{k} = (k_x, k_y)$ in the BEV image $\img{t}$, we define its descriptor $\vector{d} \in \{0, 1\}^{L^2}$ as the $L \times L$ patch of $\img{t}$ centered at $\vector{k}$, flattened into a vector. Here, $L$ is an odd number to ensure a well-defined center pixel. Patches exceeding the image border are zero-padded, consistent with the convention of Sec.~\ref{sec:correlationflow}. The descriptor is thus a binary vector encoding the occupancy pattern in the key point's neighborhood. Fig.~\ref{fig:key_pt_descriptor} illustrates its computation.

\vspace{-2mm}
\subsection{Matching keypoints}
Let $\mathcal{K}_t$ and $\mathcal{K}_{t+1}$ denote the sets of key points identified in $\img{t}$ and $\img{t+1}$, respectively.
To compute their displacement, we establish correspondences between $\mathcal{K}_t$ and $\mathcal{K}_{t+1}$.
We measure the similarity between two key points $\vector{k}_{t,i} \in \mathcal{K}_t$ and $\vector{k}_{t+1,j} \in \mathcal{K}_{t+1}$ using the normalized correlation of their descriptors:
\begin{equation}
    R(\vector{d}_{t,i}, \vector{d}_{t+1,j}) = \frac{
        \vector{d}_{t,i}^{\top} \vector{d}_{t+1,j}
    }{
        \| \vector{d}_{t,i} \| \, \| \vector{d}_{t+1,j} \|
    },
    \label{eq:descriptor_ncc}
\end{equation}
where $\| \cdot \|$ denotes the Euclidean norm.
The correspondence of $\vector{k}_{t,i}$ is the key point $\vector{k}_{t+1,j^*}$ with the descriptor most strongly correlated with $\vector{d}_{t,i}$:
\begin{equation}
    j^* = \argmax_{j} \, R(\vector{d}_{t,i}, \vector{d}_{t+1,j}).
    \label{eq:best_match}
\end{equation}

To reject false matches, we apply Lowe's ratio test \cite{lowe2004sift}: a match is accepted only if its correlation exceeds that of the second-best match by a sufficient margin: 
\begin{equation}
    R(\vector{d}_{t,i}, \vector{d}_{t+1,j^*}) > \rho \cdot R(\vector{d}_{t,i}, \vector{d}_{t+1,j^{**}})
\end{equation}
where, $j^{**}$ is the second-best match and $\rho > 1$ is the ratio threshold.

\subsection{From sparse scene flow to dense scene flow}
Under the piecewise-rigidity assumption, we densify the sparse key-point flow. We extract connected components from $\img{t}$ and assign to each point the median flow of its component's matched keypoints. This single-flow-per-component assignment assumes translational motion, consistent with the negligible inter-frame rotation established in Sec.~\ref{sec:correlationflow}. Components with no accepted matches are assigned zero flow. This step turns the sparse set of confident correspondences into a dense flow field over all points of $\pc{t}$. Note that clustering can be avoided entirely by inferring the scene flow of regular points from that of key points using K-nearest-neighbor interpolation.

\section{Experiments}
\label{sec:experiments}
We first conduct an ablation study to assess the impact of each component of CorrelationFlow, then compare our method against state-of-the-art approaches to demonstrate its effectiveness.

\textbf{Datasets.}
For our ablation study, we use the Argoverse 2 Sensor dataset (AV2)~\cite{wilson2021argoverse2}, collected using two roof-mounted 32-beam LiDARs. The dataset consists of 1,000 sequences, split into 700 for training, 150 for validation, and 150 for testing. Each sequence lasts approximately \SI{15}{s} and contains point clouds captured at 10 Hz. We compare CorrelationFlow to state-of-the-art unsupervised methods using the test set of the AV2 2026 Multi-Dataset Scene Flow Challenge~\cite{li2025uniflowzeroshotlidarscene}. This test set comprises 9,613 point clouds from five datasets: AEVA~\cite{aevascenes}, Argoverse 2~\cite{wilson2021argoverse2}, nuScenes~\cite{caesar2020nuscenes}, TruckScenes~\cite{fent2024truckscenes}, and Waymo Open Dataset~\cite{waymo}. It assesses how a single method generalizes across heterogeneous LiDAR sensors, vehicle platforms, and geographic regions without per-dataset tuning. 
All datasets provide ground-point masks, which we use to remove ground points prior to scene flow estimation.

\textbf{Metrics.}  
In all experiments, we evaluate scene flow performance using the Bucket Normalized End-Point Error (EPE)~\cite{khatri2024trackflow} for dynamic points, which we refer to as Dynamic Bucket Normalized EPE throughout this section. While the standard EPE measures the Euclidean distance between predicted and ground-truth flow vectors, it is heavily biased toward vehicles when evaluating dynamic points for two reasons. First, vehicles, due to their larger size, generate significantly more points in point clouds than pedestrians. Second, vehicles typically move faster than pedestrians. Consequently, a scene flow method that completely fails on pedestrians can still perform well on benchmarks if it excels at predicting vehicle motion.

Bucket Normalized EPE addresses this bias by first grouping points into classes, then into speed buckets (e.g., 0--0.4~m/s, 0.4--0.8~m/s). This allows normalizing the EPE of points in each bucket by their average speed, effectively transforming the evaluation from absolute error to fractional error. Thus, instead of measuring average prediction error in centimeters, Bucket Normalized EPE measures the percentage of ground-truth motion that the model fails to predict. A 0.1 improvement in this metric means predictions are on average 10\% more accurate. While this difference is negligible in low-speed buckets (e.g., 0--0.4~m/s), it translates to a 1.0~m error reduction over a 1-second interval for a car moving at 10~m/s. Object classes include regular cars (CAR), other vehicles (OTHER), pedestrians (PED), and wheeled vulnerable road users (VRU). All evaluations are conducted within a \SI{70}{m} $\times$ \SI{70}{m} area centered on the ego vehicle.

\subsection{Ablation study}

We assess four key components of our method: the necessity of an additional view to measure the Z-component of scene flow, the effectiveness of our clustering method based on connected components, the benefit of incorporating past point clouds, and the improvement achieved through the multi-scale strategy.
It should be noted that the reported execution times were obtained on a machine executing multiple threads in parallel. Consequently, the absolute timings do not represent the fastest possible execution, and our comparison instead emphasizes the relative differences between methods.

\textbf{Impact of pixel size.}
Tab.~\ref{tab:which_pixel_size} shows the evolution of scene flow error with respect to the pixel size of the BEV image of input point clouds.
We can observe that increasing the pixel size reduces the execution time.
This is explained by the reduction of the size of the BEV image and the linear complexity of the connected components algorithm \cite{bolelli2022connected} with respect to the input image size that we use to detect spatio-temporal clusters in input point clouds.
\begin{table}[b]
\resizebox{\linewidth}{!}{
    \centering
    \begin{tabular}{ c  c  c  c  c  c  c}
    \toprule
     Pixel & \multicolumn{5}{c}{Dynamic Bucket Normalized EPE $\downarrow$} & Exec.\\
     size (m) & CAR & OTHER & PED. & VRU & Mean & Time (s)\\
     
     \midrule
     
     0.05 & 0.5116 & 0.5350 & 0.4778 & 0.5371 & 0.5154 & 1.6537 \\
     
     0.1 & 0.3093 & 0.3458 & 0.3746 & 0.3799 & 0.3524 & 0.4674 \\

     0.2 & 0.2476 & 0.3008 & 0.5065 & 0.3324 & 0.3468 & 0.2237 \\

     0.4 & 0.2815 & 0.3539 & 0.8474 & 0.5750 & 0.5145 & 0.1436 \\
     
     \bottomrule
    \end{tabular}
}

    \caption{\small Effect of BEV pixel size on Dynamic Bucket Normalized EPE across dynamic object classes and execution time.}
    \vspace{-2mm}
    \label{tab:which_pixel_size}
\end{table}
A too small pixel size of \SI{0.05}{m} makes every object too fragmented to cluster precisely.
This results in a high error for every class.
As the pixel size increases from \SI{0.1}{m} to \SI{0.4}{m}, the error of medium to large object classes (e.g., CAR, and OTHER) reduces because it is easier to detect their clusters entirely.
On the other hand, the error of pedestrian increases because they occupy less pixels, thus making them harder to be clustered.
Moreover, even if their clusters are correctly identified, a reduced number of pixels increases the difficulty of scene flow estimation through NCC maximization as there are multiple maxima.
As the pixel size \SI{0.1}{m} strikes a better balance between the error of small objects (PED.) and large objects (CAR, OTHER), we set the pixel size to this value for all subsequent experiments.

\textbf{Impact of views.}
The first experiment evaluates the contribution of each view to estimation accuracy. 
As shown in Tab.~\ref{tab:which_views}, adding the sectional view to compute the Z-component of the flow yields only a marginal change in error. 
However, this additional view requires an extra NCC maximization per cluster (Alg.~\ref{alg:scene_flow_1view}), increasing the flow estimation time by $56.6\%$. 
Furthermore, most driving scenes occur in flat environments, where objects of interest (vehicles, pedestrians, and VRUs) exhibit minimal vertical motion. 
For these reasons, we retain only the BEV in the base version of CorrelationFlow, which serves as the foundation for all subsequent experiments.

\begin{table}[htbp]
\resizebox{\linewidth}{!}{
    \centering
    \begin{tabular}{ c  c  c  c  c  c  c}
    \toprule
     \multirow{2}{*}{View} & \multicolumn{5}{c}{Dynamic Bucket Normalized EPE $\downarrow$} & Flow Est.\\
      & CAR & OTHER & PED. & VRU & Mean & Time (s)\\
     
     \midrule
     
     BEV & 0.3093 & 0.3458 & 0.3746 & 0.3799 & 0.3524 & 0.1943 \\
     
     +Sectional  & \redcell{+0.0004} & \greencell{-0.0003} & \greencell{-0.0005} & \redcell{+0.0055} & \redcell{+0.0013} & \redcell{+0.1099} \\

     
     \bottomrule
    \end{tabular}
}
    \caption{\small Effect of adding a sectional projection view on Dynamic Bucket Normalized EPE across dynamic object classes and flow estimation time.}

    \label{tab:which_views}
\end{table}

\vspace{-2mm}
\textbf{Impact of the clustering method.}
Existing unsupervised methods typically use density-based algorithms like DBSCAN or HDBSCAN for spatio-temporal clustering, either during training~\cite{zhang2024seflow,khoche2025ssf,zhang2025himo,zhang2026teflow} or at inference~\cite{lin2024icpflow}. 
To evaluate our connected-component-based clustering, we replace the clustering module of CorrelationFlow with HDBSCAN while keeping the rest of the pipeline unchanged. 

\begin{table}[b]
\resizebox{\linewidth}{!}{
    \centering
    \begin{tabular}{ l  c  c  c  c  c c}
    \toprule
     Clustering & \multicolumn{5}{c}{Dynamic Bucket-Normalized EPE $\downarrow$} & Clustering \\
     Method & CAR & OTHER & PED. & VRU & Mean & Time (s)\\
     
     \midrule
     
     Connected & \multirow{3}{*}{0.3093} & \multirow{3}{*}{0.3458} & \multirow{3}{*}{0.3746} &\multirow{3}{*}{0.3799} & \multirow{3}{*}{0.3524} & \multirow{3}{*}{0.2731}\\
     components & & & & & \\
      (ours) & & & & & \\

     HDBSCAN  & \greencell{0.2304} & \redcell{0.3853} & \greencell{0.3669} & \greencell{0.3161} & \greencell{0.3247} & \redcell{2.7521} \\
     
     \bottomrule
    \end{tabular}
}
    \caption{\small Comparison of connected-component clustering and HDBSCAN in terms of Dynamic Bucket Normalized EPE across dynamic object classes and clustering time.}
    \vspace{-2mm}

    \label{tab:which_clustering_method}
\end{table}

As shown in Tab.~\ref{tab:which_clustering_method}, HDBSCAN reduces the mean scene flow error by $7.8\%$ (from 0.3524 to 0.3247) but increases clustering time by a factor of 10. 
However, as we demonstrate in later experiments, the performance of CorrelationFlow with our clustering method improves significantly when incorporating past point clouds, without increasing clustering time. 
This is because our clustering algorithm, which uses the connected components method of \cite{bolelli2022connected}, has a constant complexity with respect to the number of points, as it depends only on the size of the BEV image. 
In contrast, HDBSCAN~\cite{mcinnes2017accelerated} has a complexity of $O(N\log N)$, where $N$ is the number of points. 
Consequently, our method scales better and is more suitable for optimization toward real-time performance.
We want to notice that the execution time is measured on a machine that evaluates multiple threads in parallel, so the absolute execution time does not represent the fastest possible execution of the algorithm. 
Instead, we focus on the comparison of the execution time between two clustering methods.

\begin{figure}[htbp]
    \centering
    \begin{tikzpicture}
\begin{axis}[
    width=0.95\linewidth,
    height=0.55\linewidth,   
    xlabel={Number of past point clouds},
    xlabel style={font=\footnotesize, yshift=3pt},
    ylabel={Dynamic Bucket Normalized EPE},
    ylabel style={font=\footnotesize},
    xmin=0, xmax=9,
    ymin=0.05, ymax=0.4, 
    xtick={0,1,2,3,4,5,6,7,8,9},
    ytick=
    {0,05, 0.1,0.15,0.2,0.25,0.3,0.35,0.4},
    legend pos=south west,
    legend style={font=\tiny},
    ymajorgrids=true,
    grid style=dashed,
]

\addplot[
    color=blue,
    mark=*,
    ]
    coordinates {
    (0, 0.310142)(1, 0.273596)(2, 0.232691)(3, 0.226889)(4, 0.222292)(5, 0.224237)(6, 0.226423)(7, 0.229468)(8, 0.233804)(9, 0.236825)
    };
    \addlegendentry{Car}

\addplot[
    color=green,
    mark=triangle*,
    ]
    coordinates {
    (0, 0.375453)(1, 0.344337)(2, 0.334977)(3, 0.304421)(4, 0.297913)(5, 0.30575)(6, 0.30467)(7, 0.303267)(8, 0.302959)(9, 0.306942)
    };
    \addlegendentry{Pedestrian}

\addplot[
    color=orange,
    mark=diamond*,
    ]
    coordinates {
    (0, 0.380067)(1, 0.345536)(2, 0.33224)(3, 0.336261)(4, 0.33241)(5, 0.332962)(6, 0.335052)(7, 0.341618)(8, 0.348806)(9, 0.362223)
    };
    \addlegendentry{VRU}

\addplot[
    color=red,
    mark=square*,
    ]
    coordinates {
    (0, 0.348839)(1, 0.321113)(2, 0.262283)(3, 0.249076)(4, 0.237962)(5, 0.24056)(6, 0.237203)(7, 0.235732)(8, 0.235873)(9, 0.235991)
    };
    \addlegendentry{Other}

\end{axis}
\end{tikzpicture}

\caption{\small Mean Dynamic Bucket Normalized EPE (lower is better) as a function of the number of past point clouds $h$, per object class, on the Argoverse 2 validation split. The error decreases for every class, with the largest gains between $h = 0$ and $h = 3$.}

\label{fig:benefit_historical}
\end{figure}
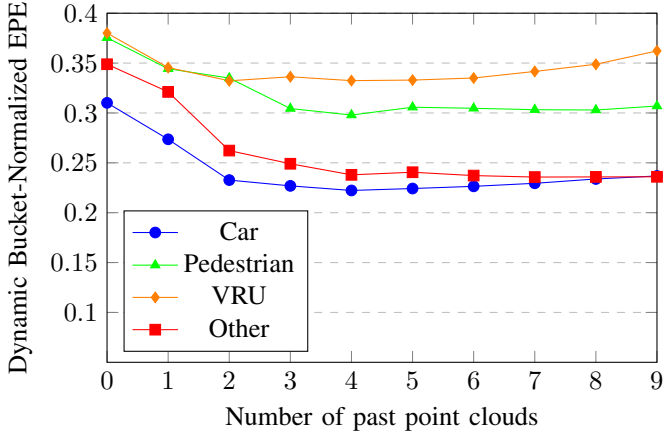

\textbf{Impact of past point clouds.}
We evaluate the benefit of using past point clouds as described in Sec.~\ref{sec:leverage_past_pcs}. 
Scene flow is estimated for every pair of point clouds collected at $(t-k)$ and $(t+1)$, with $k$ ranging from $0$ to $h$.
The results are then pooled to produce the final estimate. 
In this experiment, we use the base version of CorrelationFlow and vary $h$ from $0$ to $8$. Fig.~\ref{fig:benefit_historical} shows the mean error as a function of $h$. 
Increasing $h$ reduces the error for all object classes, with the largest reduction occurring between $h=0$ (the minimum required for flow estimation) and $h=2$; beyond $h=2$, the gains diminish.
We attribute this improvement to two effects. 
First, aggregating more past point clouds makes small and slow-moving objects, such as pedestrians, more visible in the BEV.
As a result, it is less difficult to precisely identify their spatio-temporal cluster.
Second, additional past point clouds provide multiple estimates of the scene flow $\vector{f}_{t \rightarrow t+1}$, and pooling these estimates reduces the overall error.

\begin{figure}[t!]
\centering
\begin{tikzpicture}
\begin{groupplot}[
    group style={
        group size=2 by 2, 
        vertical sep=1.5 cm, 
        horizontal sep=1.cm,
    },
    width=0.36\linewidth, 
    height=0.18\textwidth, 
    scale only axis,
    ylabel style={font=\small},
    xlabel style={font=\small},
    tick label style={font=\scriptsize},
    legend style={font=\tiny},
]

\nextgroupplot[
    xlabel={Number of scales},
    ylabel={Dynamic Bucket Normalized EPE},
    xlabel style={font=\footnotesize, yshift=5pt},
    ylabel style={font=\footnotesize},
    xmin=1, xmax=5,
    ymin=0, ymax=0.4,
    xtick={1,2,3,4,5},
    ytick={0, 0.1,0.15,0.2,0.25,0.3,0.35,0.4},
    legend pos=south west,
    ymajorgrids=true,
    grid style=dashed,
    title={\small Car},
]
\addplot[
    color=blue,
    mark=*,
    ]
    coordinates {
    (1, 0.310142)(2, 0.229121)(3, 0.198296)(4, 0.191286)(5, 0.191767)
    };
    \addlegendentry{$h=0$}

\addplot[
    color=red,
    mark=square*,
    ]
    coordinates {
    (1, 0.232691)(2, 0.196965)(3, 0.183765)(4,0.181743)(5, 0.157336)
    };
    \addlegendentry{$h=2$}

\addplot[
    color=green,
    mark=triangle*,
    ]
    coordinates {
    (1,0.222292)(2,0.192773)(3,0.179371)(4,0.151362)(5,0.159419)
    };
    \addlegendentry{$h=4$}

\nextgroupplot[
    xlabel={Number of scales},
    xlabel style={font=\footnotesize, yshift=5pt},
    ylabel={},
    xmin=1, xmax=5,
    ymin=0, ymax=0.4,
    xtick={1,2,3,4,5},
    ytick={0,0.1,0.15,0.2,0.25,0.3,0.35,0.4},
    legend pos=south west,
    ymajorgrids=true,
    grid style=dashed,
    title={\small Pedestrian},
]
\addplot[
    color=blue,
    mark=*,
    ]
    coordinates {
    (1, 0.375453)(2, 0.333332)(3, 0.31663)(4, 0.316773)(5, 0.324084)
    };
    \addlegendentry{$h=0$}

\addplot[
    color=red,
    mark=square*,
    ]
    coordinates {
    (1, 0.334977)(2, 0.349634)(3, 0.337105)(4, 0.337322)(5, 0.281792)
    };
    \addlegendentry{$h=2$}

\addplot[
    color=green,
    mark=triangle*,
    ]
    coordinates {
    (1, 0.297913)(2, 0.316727)(3, 0.311005)(4, 0.245921)(5, 0.246078)
    };
    \addlegendentry{$h=4$}

\nextgroupplot[
    xlabel={Number of scales},
    ylabel={Dynamic Bucket Normalized EPE},
    xlabel style={font=\footnotesize, yshift=5pt},
    ylabel style={font=\footnotesize},
    xmin=1, xmax=5,
    ymin=0, ymax=0.4,
    xtick={1,2,3,4,5},
    ytick={0, 0.1,0.15,0.2,0.25,0.3,0.35,0.4},
    legend pos=south west,
    ymajorgrids=true,
    grid style=dashed,
    title={\small VRU},
]
\addplot[
    color=blue,
    mark=*,
    ]
    coordinates {
    (1, 0.380067)(2, 0.26472)(3, 0.264545)(4, 0.264365)(5, 0.270231)
    };
    \addlegendentry{$h=0$}

\addplot[
    color=red,
    mark=square*,
    ]
    coordinates {
    (1, 0.33224)(2, 0.26485)(3, 0.262539)(4, 0.263389)(5, 0.238388)
    };
    \addlegendentry{$h=2$}

\addplot[
    color=green,
    mark=triangle*,
    ]
    coordinates {
    (1, 0.33241)(2, 0.259681)(3, 0.26275)(4, 0.240299)(5, 0.248639)
    };
    \addlegendentry{$h=4$}

\nextgroupplot[
    xlabel={Number of scales},
    xlabel style={font=\footnotesize, yshift=5pt},
    ylabel={},
    xmin=1, xmax=5,
    ymin=0, ymax=0.4,
    xtick={1,2,3,4,5},
    ytick={0,0.1,0.15,0.2,0.25,0.3,0.35,0.4},
    legend pos=south west,
    ymajorgrids=true,
    grid style=dashed,
    title={\small Other},
]
\addplot[
    color=blue,
    mark=*,
    ]
    coordinates {
    (1, 0.348839)(2, 0.262287)(3, 0.228462)(4, 0.204802)(5, 0.198244)
    };
    \addlegendentry{$h=0$}

\addplot[
    color=red,
    mark=square*,
    ]
    coordinates {
    (1, 0.262283)(2, 0.211254)(3, 0.1869)(4, 0.177282)(5, 0.163808)
    };
    \addlegendentry{$h=2$}

\addplot[
    color=green,
    mark=triangle*,
    ]
    coordinates {
    (1, 0.237962)(2, 0.190857)(3, 0.176007)(4, 0.155654)(5, 0.154726)
    };
    \addlegendentry{$h=4$}

\end{groupplot}
\end{tikzpicture}
\caption{\small Mean Dynamic Bucket Normalized EPE (lower is better) as a function of the number of scales in the coarse-to-fine strategy. Each panel shows one object class; each line corresponds to a number of past point clouds $h \in \{0, 2, 4\}$.}
\label{fig:benefit_multiscale}
\end{figure}
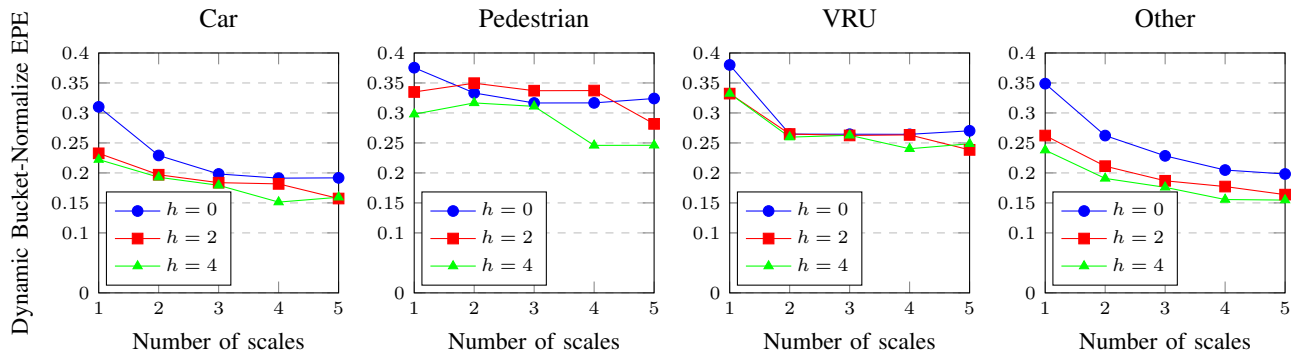

\textbf{Impact of the multi-scale strategy.}
Building on the previous experiment where the base model was enabled to use past point clouds, we now evaluate the benefit of the coarse-to-fine strategy described in Sec.~\ref{sec:correaltion_flow_components}.
Fig.~\ref{fig:benefit_multiscale} reports the mean error per object class as the number of scales varies from 1 to 5, for $h \in \{0, 2, 4\}$.
These three values of $h$ are chosen based on the previous experiment: $h = 0$ is the minimum required for flow estimation, the largest improvement occurs between $h = 0$ and $h = 2$, and the best performance is achieved at $h = 4$.
For all classes, the lowest error is consistently achieved with $h=4$ at every scale. 
With $h=0$, increasing the number of scales reduces the error, with the most significant reduction occurring when the number of scales increases from $1$ to $2$. 
The rate of reduction diminishes as the number of scales exceeds $3$. 
Compared to $h=0$, the error reduction with higher values of $h$ (e.g., $2$ or $4$) is smaller as the number of scales increases. 
A common trend across all values of $h$ is that the error for the Other class, which comprises large vehicles, decreases linearly with the number of scales.
Such a decrease indicates the effectiveness of the multi-scale strategy in identifying the entirety of large objects.

\begin{table}[b!]

    \centering
    \begin{tabular}{ p{1.75cm}  c  c  c  c  c }
    \toprule
     \multirow{2}{*}{Method} & \multicolumn{5}{c}{Dynamic Bucket Normalized EPE $\downarrow$} \\
      & CAR & OTHER & PED. & VRU & Mean \\
     
     \midrule
     
     Connected components & \textbf{0.1701} & \textbf{0.2633} & \textbf{0.4401} & \textbf{0.2647} & \textbf{0.2845} \\
     
     Nearest neighbor  & 0.3398 & 0.5845 & 0.8286 & 0.4519 & 0.5512 \\

     \textcolor{gray}{CorrelationFlow ($h=0$)} & \textcolor{gray}{0.3093} & \textcolor{gray}{0.3458} & \textcolor{gray}{0.3746} & \textcolor{gray}{0.3799} & \textcolor{gray}{0.3524} \\

     \textcolor{gray}{CorrelationFlow ($h=4$)} & \textcolor{gray}{0.2269} & \textcolor{gray}{0.2491} & \textcolor{gray}{0.3044} & \textcolor{gray}{0.3363} & \textcolor{gray}{0.2792} \\
     
     \bottomrule
    \end{tabular}


    \caption{\small Effect of sparse-to-dense flow densification on CorrelationFlow-Keypoints using connected-component and nearest-neighbor assignments. Gray rows report the results of CorrelationFlow for references. The better results are highlighted in bold.}
    \label{tab:keypoint_sparse2dense}
\end{table}

\textbf{Necessity of clustering -- CorrelationFlow-Keypoints.}
\begin{figure*}[ht!]
\centering
\begin{tikzpicture}
\begin{axis}[
    width=\linewidth,
    height=0.25\linewidth, 
    ybar,                          
    bar width=0.12cm,              
    enlarge x limits=0.15,        
    legend style={at={(0.5,1.05)}, anchor=south, legend columns=8}, 
    ylabel={Dynamic Bucket Normalized EPE},                
    xlabel={},              
    symbolic x coords={Overall, AEVA, AV2, nuScenes, TruckScene, Waymo},
    xtick=data,                    
    ymin=0.1, ymax=0.9,                
    ytick={0.1,0.3,0.5,0.7,0.9},   
    nodes near coords align={vertical},
    x tick label style={rotate=0, anchor= north }, 
    xlabel style={font=\footnotesize},
    ylabel style={font=\footnotesize},
    legend style={at={(0.5,1.05)},
    anchor=south,
    legend columns=9, 
    font=\footnotesize},
]

\definecolor{color0}{RGB}{0,0,255}      
\definecolor{color1}{RGB}{255,0,0}      
\definecolor{color2}{RGB}{0,200,0}      
\definecolor{color3}{RGB}{255,165,0}    
\definecolor{color4}{RGB}{128,0,128}    
\definecolor{color5}{RGB}{255,192,203}  
\definecolor{color6}{RGB}{0,128,128}    
\definecolor{color7}{RGB}{75,0,130}    
\definecolor{color8}{RGB}{165,42,42}    

\addplot[color=color6, fill=color8] coordinates {(Overall, 0.1899) (AEVA, 0.2720) (AV2, 0.1279) (nuScenes, 0.1784) (TruckScene, 0.2523) (Waymo, 0.1190)};

\addplot[color=color0, fill=color0] coordinates {(Overall, 0.2668) (AEVA, 0.3603) (AV2, 0.2106) (nuScenes, 0.2445) (TruckScene, 0.3043) (Waymo, 0.2141)};

\addplot[color=color2, fill=color2] coordinates {(Overall, 0.2720) (AEVA, 0.3273) (AV2, 0.1632) (nuScenes, 0.3702) (TruckScene, 0.3183) (Waymo, 0.1813)};

\addplot[color=color3, fill=color3] coordinates {(Overall, 0.3116) (AEVA, 0.3875) (AV2, 0.2019) (nuScenes, 0.3822) (TruckScene, 0.3884) (Waymo, 0.1979)};

\addplot[color=color7, fill=color7] coordinates {(Overall, 0.4092) (AEVA, 0.4636) (AV2, 0.2952) (nuScenes, 0.3844) (TruckScene, 0.6273) (Waymo, 0.2756)};

\addplot[color=color1, fill=color1] coordinates {(Overall, 0.4133) (AEVA, 0.6300) (AV2, 0.1597) (nuScenes, 0.4338) (TruckScene, 0.6409) (Waymo, 0.2020)};

\addplot[color=color5, fill=color5] coordinates {(Overall, 0.5855) (AEVA, 0.7961) (AV2, 0.2904) (nuScenes, 0.7035) (TruckScene, 0.8310) (Waymo, 0.3067)};

\addplot[color=color4, fill=color4] coordinates {(Overall, 0.5919) (AEVA, 0.8057) (AV2, 0.2827) (nuScenes, 0.6965) (TruckScene, 0.8555) (Waymo, 0.3192)};

\addplot[color=color6, fill=color6] coordinates {(Overall, 0.6289) (AEVA, 0.8412) (AV2, 0.3113) (nuScenes, 0.7543) (TruckScene, 0.8885) (Waymo, 0.3494)};

\legend{SynFlow, CorrelationFlow, RVLoss, TeFlow, CorrelationFlow-Keypoints, SSF, VoteFlow, SeFlow++, SeFlow}

\end{axis}
\end{tikzpicture}
\caption{\small Comparison against state-of-the-art unsupervised methods on the test split of the Argoverse 2 2026 Multi-Dataset Scene Flow Challenge, measured by the mean Dynamic Bucket Normalized EPE (lower is better) over the short (0--\SI{35}{m}) range. On each dataset, this metric is averaged over four object classes (car, other vehicles, pedestrians, and VRU). 
} 
\label{fig:experiment_av2challenge_across_dataset}
\end{figure*}
We compare the keypoint-based variant of CorrelationFlow against the original version to assess whether explicit spatio-temporal clustering is necessary for accurate scene flow estimation. Recall that CorrelationFlow-Keypoints operates on a single sweep pair and therefore does not exploit past point clouds; thus, the relevant baseline is CorrelationFlow at $h=0$. The results in Tab.~\ref{tab:keypoint_sparse2dense} show that, when the sparse keypoint flow is densified using connected components, the keypoint variant achieves a mean Dynamic Bucket Normalized EPE of $0.2845$, improving over the dense CorrelationFlow at $h=0$ ($0.3524$) by $19.3\%$ and nearly matching its best configuration at $h=4$ ($0.2792$), despite relying on a single sweep pair rather than a temporal window. The gain is consistent across the CAR ($0.1701$ vs.\ $0.3093$), OTHER ($0.2633$ vs.\ $0.3458$), and VRU ($0.2647$ vs.\ $0.3799$) classes; the pedestrian class is the sole exception ($0.4401$ vs.\ $0.3746$), as thin, sparsely sampled pedestrian footprints provide few repeatable boundary keypoints and thus fewer confident matches.

These results indicate that explicit clustering is not a prerequisite for competitive scene flow: matching lightweight boundary descriptors on a single sweep pair recovers most of the accuracy that the dense variant obtains only after aggregating several past sweeps. However, the densification strategy is decisive. Replacing the connected-component densification with a nearest-neighbor assignment degrades the mean EPE from $0.2845$ to $0.5512$, nearly doubling the error in every class. This confirms that propagating the sparse keypoint flow under the piecewise-rigidity prior, assigning each connected component the median flow of the matched keypoints it contains, is essential to turning reliable but sparse correspondences into an accurate dense flow field, whereas a purely geometric nearest-neighbor densification discards the rigidity structure and propagates matching errors.


\subsection{Comparison to other unsupervised methods}

We compare CorrelationFlow and its key-point-based variant, CorrelationFlow-Keypoints, against state-of-the-art unsupervised methods on the test split of the Argoverse 2 2026 Multi-Dataset Scene Flow Challenge.
As described in Sec.~\ref{sec:experiments}, this test set contains 9{,}613 point clouds drawn from five source datasets (AEVA, Argoverse 2, nuScenes, TruckScenes, and Waymo Open Dataset), spanning heterogeneous LiDAR sensors, vehicle platforms, and geographic regions; a method must therefore generalize across domains rather than fit a single sensor configuration, which is precisely the setting where a training-free approach is expected to shine.
Our comparison includes six baselines: SeFlow \cite{zhang2024seflow,zhang2025himo}, a self-supervised feed-forward network trained with cycle-consistency objectives on a PointPillars-style backbone; SeFlow++ \cite{zhang2025himo}, its improved successor; TeFlow \cite{zhang2026teflow}, which adds temporal ensembling across multiple frames; VoteFlow \cite{lin2025voteflow}, which embeds a rigid-motion prior in the backbone; SSF \cite{khoche2025ssf}, which targets sparse long-range scene flow; and RVLoss \cite{wang2026rvloss}.
All baselines are self-supervised: they require no manual annotations but rely on large-scale training, whereas our methods require neither training nor labels.
We include SynFlow \cite{zhang2026synflow} for completeness; however, although it does not rely on manual annotations, it is fully supervised using large-scale synthetic data. Its performance therefore reflects the scale and quality of simulated training data rather than advances in unsupervised scene flow estimation.

The result in Fig.~\ref{fig:experiment_av2challenge_across_dataset} shows that CorrelationFlow achieves that best overall performance, outperforming state-of-the-art data-driven unsupervised methods.
We can observe that every method transfer well from AV2 to Waymo as they have similar point cloud density and LiDAR range, making this transfer in-domain.
Compared to these two datasets, nuScenes is a different domain as its LiDAR has only half the number of ray.
As a result, data-driven methods transfer suffer a large performance drop in this dataset.
CorrelationFlow, on the other hand, keeps the same performance as its Dynamic Bucket Normalized EPE only change by 0.03 from AV2 (or Waymo) to nuScenes.
Such a performance is a strong indicator of the generalisability of CorrelationFlow.
The generalisation to AEVA and TruckScene is more challenging as the former offers an entirely different LiDAR with a completely different scanning pattern and latter contains data in highway with objects moving with high-speed, posing more difficulty to the connected-component-based clustering algorithm.


\subsection{Runtime Analysis}
We evaluate the runtime of CorrelationFlow-Keypoints on the AV2 validation set. The model processes points within a 51.2\,m range, corresponding to a $102.4 \times 102.4\,\text{m}^2$ region of interest in the horizontal plane. This region is larger than the $76.8 \times 76.8\,\text{m}^2$ region of interest considered in existing works~\cite{zhang2024deflow, zhang2026teflow}. We evaluate two implementations of CorrelationFlow-Keypoints: the original CPU implementation, executed on an AMD EPYC 9534 processor using eight threads, and a PyTorch implementation executed on an NVIDIA A100 80\,GB GPU. The runtimes of both implementations are reported in Tab.~\ref{tab:correlationflow_runtime}.


\begin{table}[h!]
    \centering
    \small
    \setlength{\tabcolsep}{2pt}

    \begin{threeparttable}
        \caption{\small Runtime analysis of CorrelationFlow-KeyPoints on
        the validation split of Argoverse 2.}
        \label{tab:correlationflow_runtime}

        \begin{tabularx}{\columnwidth}{@{}Xrrr@{}}
            \toprule
            Component
                & \multicolumn{1}{c}{\mbox{CPU (ms)}}
                & \multicolumn{1}{c}{\mbox{GPU (ms)}}
                & \multicolumn{1}{c}{\mbox{Speedup ($\times$)}} \\
            \midrule

            Ego-motion Compensation
            & 3.09 & 0.34 & 9.11 \\
        BEV Projection
            & 6.46 & 0.47 & 13.72 \\
        Keypoint Detection
            & 6.19 & 1.39 & 4.44 \\
        Descriptor Construction
            & 305.38 & 0.52 & 591.54 \\
        Keypoint Matching
            & 1120.38 & 29.94 & 37.42 \\
        Flow Propagation
            & 34.29 & 7.41 & 4.63 \\
        \midrule
        Measured Component Subtotal
            & 1475.79 & 40.07 & 36.83 \\

            Other Inference Overhead\tnote{$\dagger$}
                & 2.87 & 0.32 & -- \\

            \textbf{Total Inference}
                & \textbf{1478.66}
                & \textbf{40.39}
                & \textbf{36.61} \\
            \bottomrule
        \end{tabularx}

        \begin{tablenotes}[flushleft]
            \footnotesize
            \item[$\dagger$] Other inference overhead is the residual
            between the synchronized forward latency and the measured
            component subtotal. CPU components use synchronous
            host-wall intervals, whereas GPU components use CUDA-event
            intervals. Data loading and metric computation are excluded.
        \end{tablenotes}
    \end{threeparttable}
\end{table}

\begin{figure*}[htbp]
    \centering
    \setlength{\tabcolsep}{1pt}
    \renewcommand{\arraystretch}{1.1}
    \newcommand{\qw}{0.187\linewidth}
    \newcommand{\colhead}[1]{\shortstack{\footnotesize #1}}
    \newcommand{\rowlab}[1]{\rotatebox[origin=c]{90}{\footnotesize #1}}
    \begin{tabular}{@{}c@{\hspace{1pt}}ccccc@{}}
        & \colhead{\footnotesize Ego motion\\ \footnotesize compensated}
        & \colhead{\footnotesize Correlation\\ \footnotesize Flow}
        & \colhead{\footnotesize Correlation\\ \footnotesize Flow-KeyPoints}
        & \colhead{\footnotesize SSF}
        & \colhead{\footnotesize TeFlow} \\[2pt]

        \rowlab{Truck}
        & \includegraphics[width=\qw]{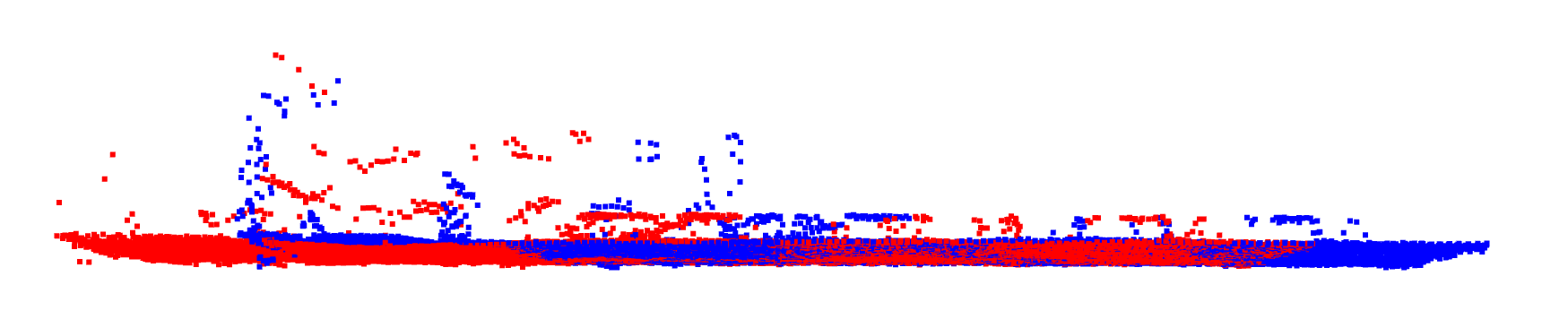}
        & \includegraphics[width=\qw]{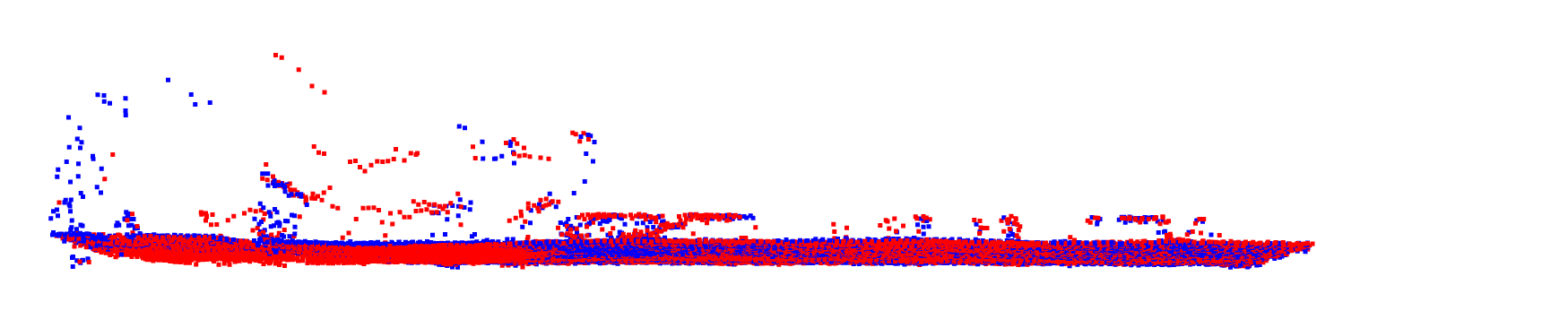}
        & \includegraphics[width=\qw]{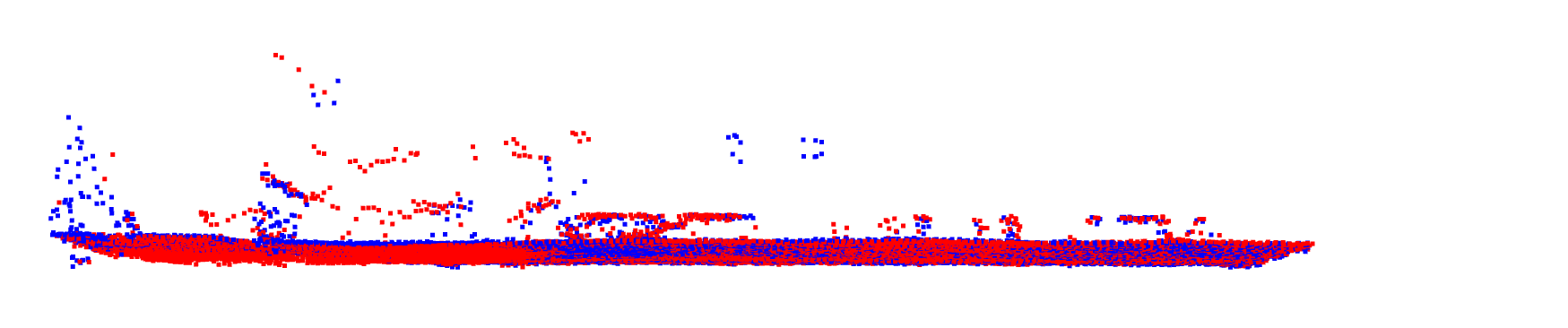}
        & \includegraphics[width=\qw]{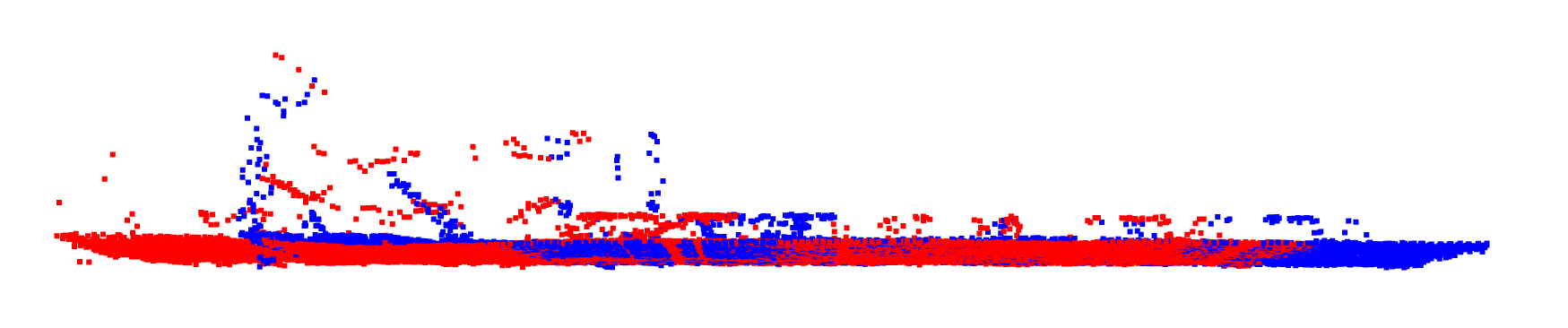}
        & \includegraphics[width=\qw]{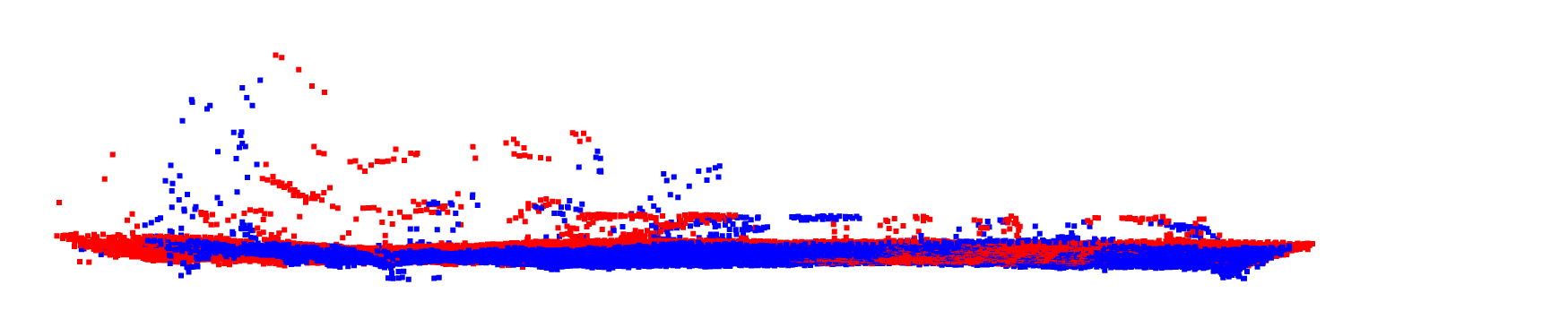} \\[2pt]

        \rowlab{\shortstack{\footnotesize Close \\ \footnotesize range}}
        & \includegraphics[width=\qw]{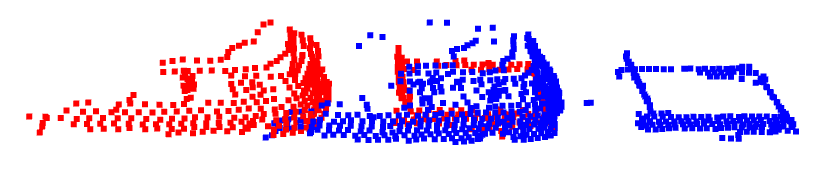}
        & \includegraphics[width=\qw]{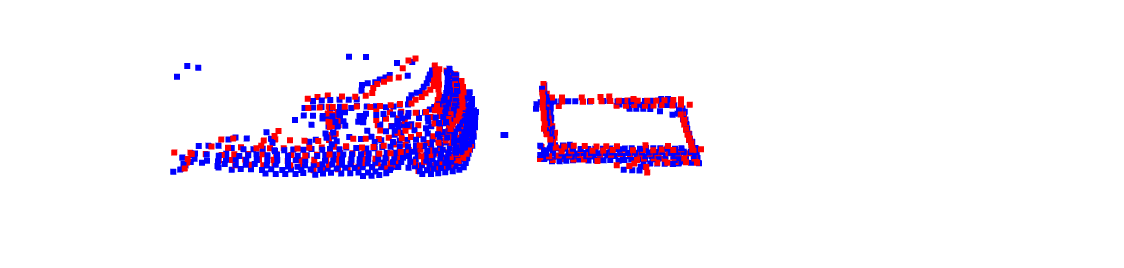}
        & \includegraphics[width=\qw]{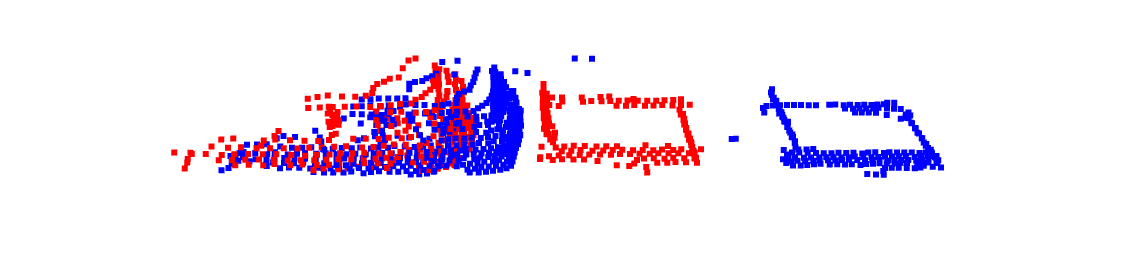}
        & \includegraphics[width=\qw]{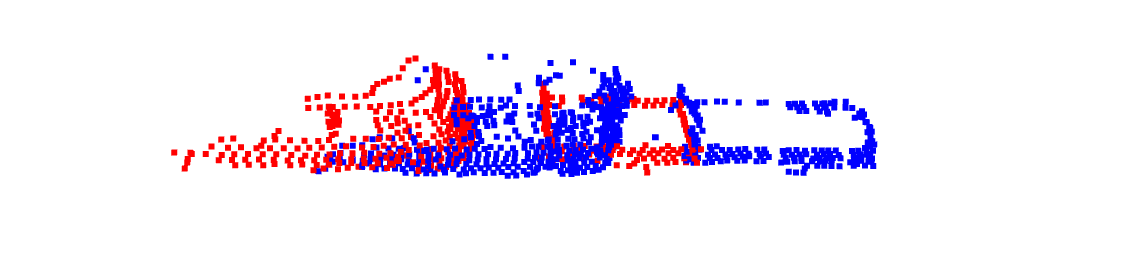}
        & \includegraphics[width=\qw]{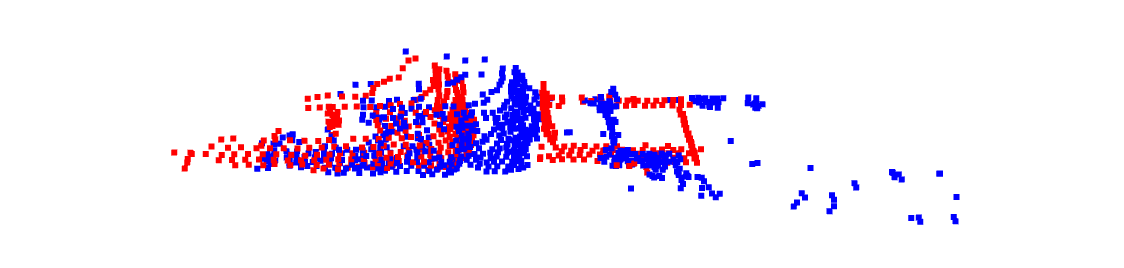} \\[2pt]

        \rowlab{\shortstack{\footnotesize Typical\\ \footnotesize scene}}
        & \includegraphics[width=\qw]{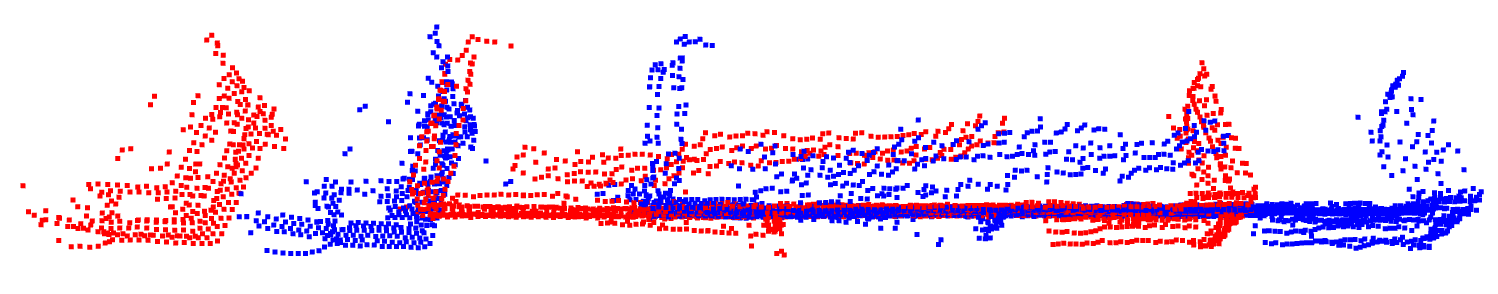}
        & \includegraphics[width=\qw]{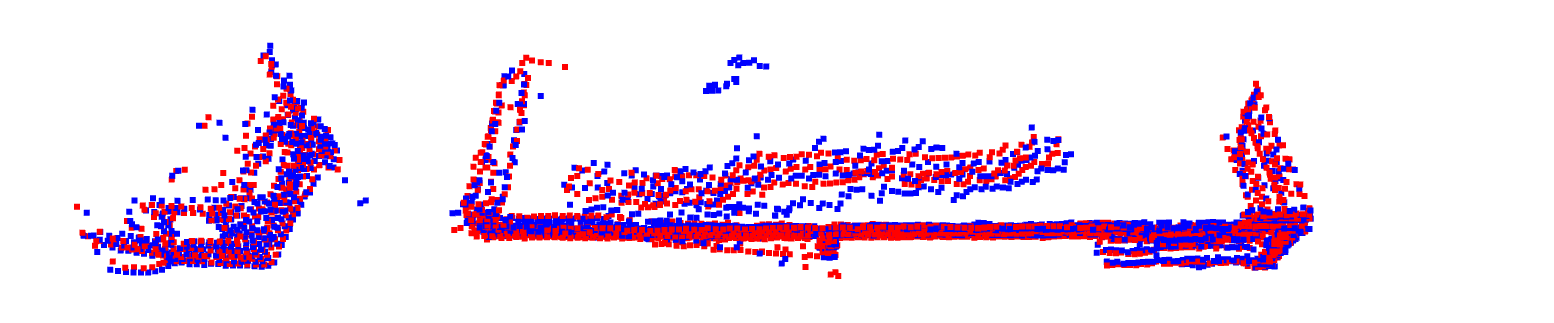}
        & \includegraphics[width=\qw]{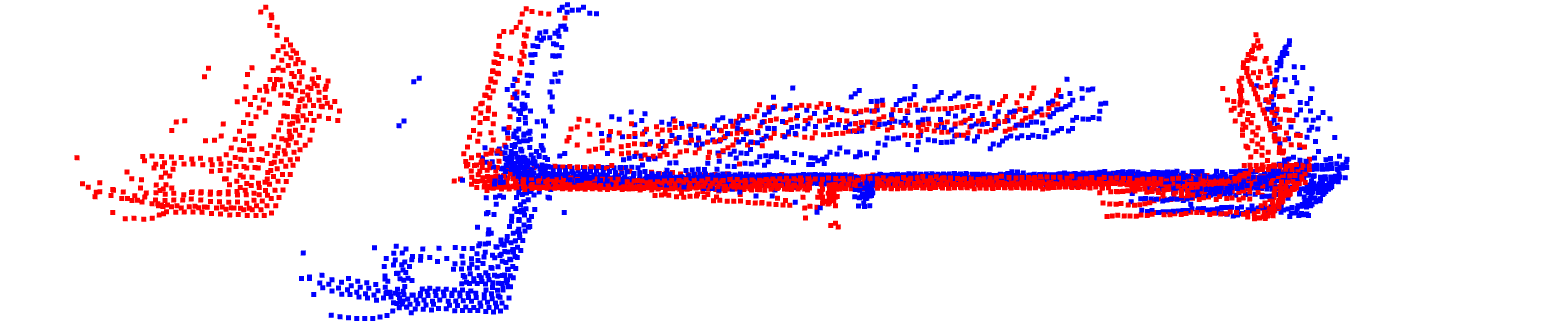}
        & \includegraphics[width=\qw]{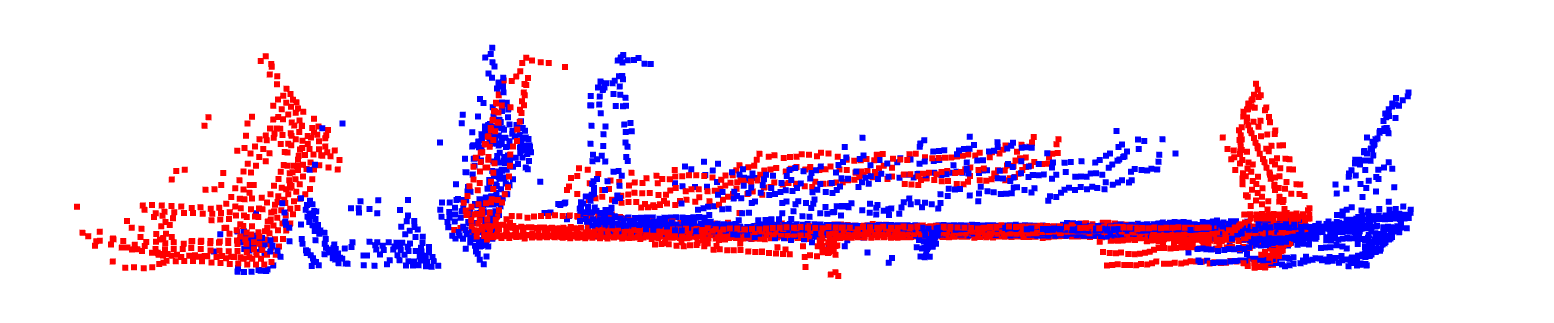}
        & \includegraphics[width=\qw]{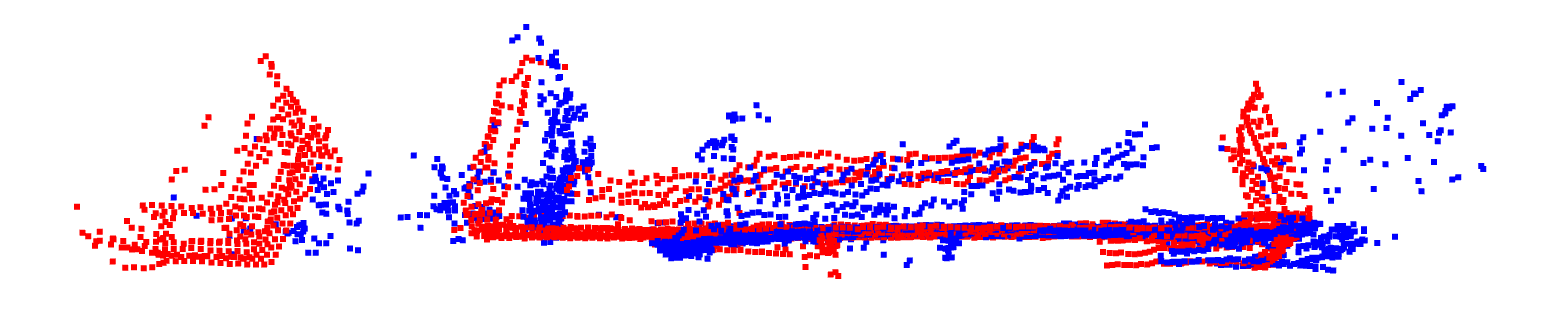} \\[2pt]

        \rowlab{\shortstack{\footnotesize Different\\ \footnotesize LiDAR}}
        & \includegraphics[width=\qw]{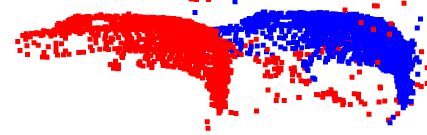}
        & \includegraphics[trim=0cm 1cm 0cm 0cm, clip=true, width=\qw]{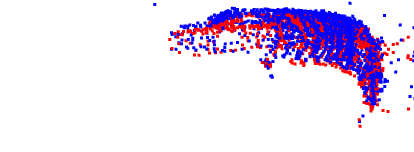}
        & \includegraphics[trim=0cm 1cm 0cm 0cm, clip=true, width=\qw]{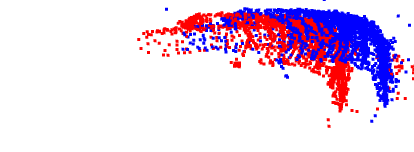}
        & \includegraphics[trim=0cm 1cm 0cm 0cm, clip=true, width=\qw]{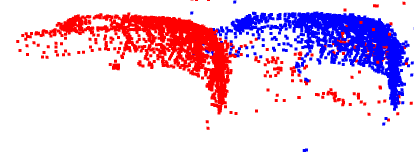}
        & \includegraphics[trim=0cm 1cm 0cm 0cm, clip=true, width=\qw]{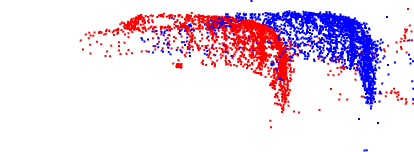} \\[2pt]

        \rowlab{\shortstack{\footnotesize Pedestrians}}
        & \includegraphics[trim=0cm 2cm 0cm 1cm, clip=true,width=\qw]{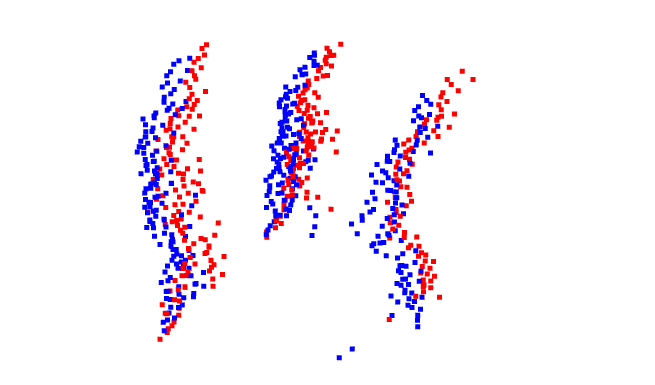}
        & \includegraphics[trim=0cm 2cm 0cm 1cm, clip=true,width=\qw]{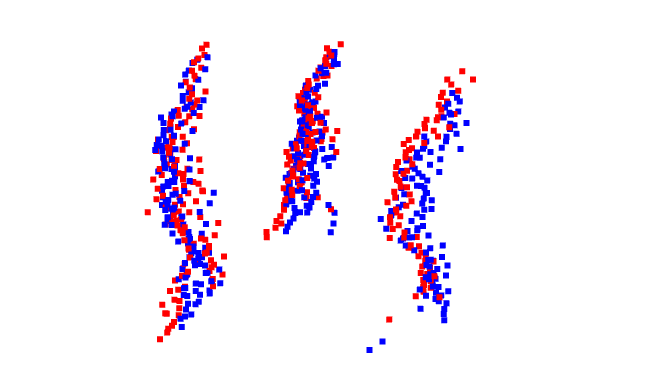}
        & \includegraphics[trim=0cm 2cm 0cm 1cm, clip=true,width=\qw]{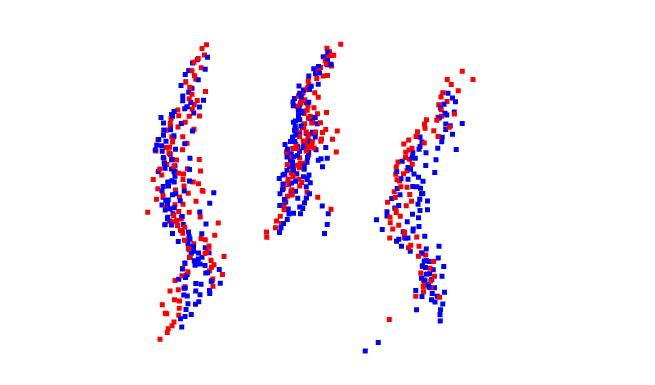}
        & \includegraphics[trim=0cm 2cm 0cm 1cm, clip=true,width=\qw]{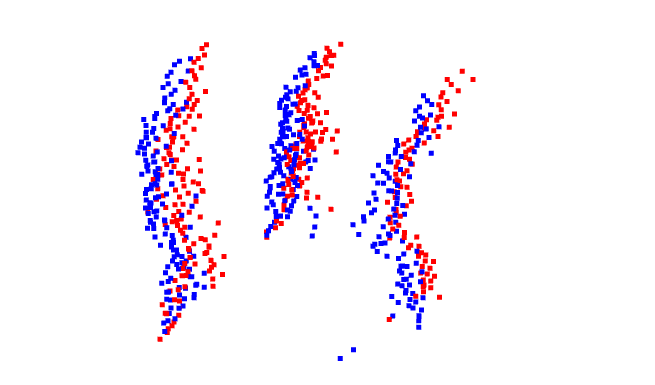}
        & \includegraphics[trim=0cm 2cm 0cm 1cm, clip=true,width=\qw]{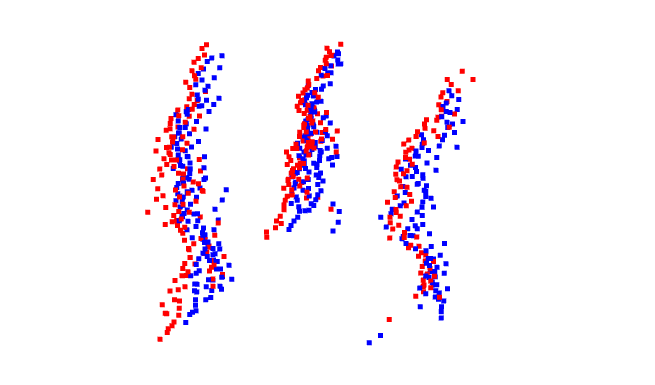} \\
    \end{tabular}

    \caption{\small Qualitative results on the test set of the Argoverse~2
    2026 Multi-Dataset Scene Flow Challenge. Each row is a different scene; each
    column a different method. The first column shows the ego-motion-compensated
    input, with the source sweep at time $t$ in blue and the target sweep at
    $(t{+}1)$ in red.
    The remaining columns overlay the source sweep after warping by each method's estimated flow onto the target: better estimates collapse the blue--red gap. Rows span a large truck, a  close-range object, a typical scene, a sequence captured with a LiDAR different from the one used in AV2, and a scene with pedestrian in close range.}
    \label{fig:qualitative_result}
\end{figure*}

The CPU implementation requires 1.48\,s per point-cloud pair, corresponding to 0.68 pairs/s, which is suitable for offline, training-free processing. The PyTorch implementation reduces the inference latency to 40.39\,ms, yielding a $36.61\times$ speedup. Keypoint matching remains the primary bottleneck, accounting for 74.13\% of the total GPU inference latency. In particular, descriptor construction requires 305.38\,ms on the CPU but only 0.52\,ms on the GPU, corresponding to a $591.54\times$ speedup. This improvement is primarily attributable to replacing costly per-keypoint NumPy operations with parallel PyTorch tensor operations that remain resident on the GPU.

AV2 LiDAR sweeps are acquired at 10\,Hz, which provides a 100\,ms processing budget for each consecutive point-cloud pair. On the GPU, CorrelationFlow-Keypoints achieves an inference latency of 40.4\,ms, equivalent to 24.8 pairs/s and $2.48\times$ the native AV2 acquisition rate. Thus, the method satisfies the real-time processing requirement while remaining entirely training-free.

\subsection{Qualitative results}
Fig.~\ref{fig:qualitative_result} shows representative predictions on the challenge test set. 
In the ego-motion-compensated input (first column), overlapping static structure appears purple, while a dynamic object leaves a blue--red gap proportional to its motion; a good estimate warps the source sweep so that this gap collapses onto the target. 
On the moving objects, the truck, the fast-moving object captured with a LiDAR different from AV2, CorrelationFlow collapses this gap most completely, whereas SSF and TeFlow leave a visible residual offset between the two sweeps. 
The keypoint variant recovers most of this motion from a single sweep pair, though its sparser correspondences leave a slightly larger residual on thin structures.
Thanks to the model-based nature of CorrelationFlow, we can explain with certainty why it performs better in these cases.
The first reason is that our connected-component-based clustering algorithm successfully identify points of these objects in their entirety despite gaps between their parts.
The second reason is that the maximization of the NCC between BEV images of instances of an object can precisely compute its scene flow.


\section{Limitations}
\label{sec:limitations}
CorrelationFlow inherits the consequences of its assumptions, as stated in Secs.~\ref{sec:correlationflow} and \ref{sec:correaltion_flow_components}.
First, the BEV projection in the base version discards vertical motion and conflates points at different heights within a pixel; the sectional view recovers the vertical component, but at additional cost and only under the planar-motion assumption.
Second, the piecewise-rigidity model ties the accuracy to the quality of the clustering: under-segmentation, i.e., nearby objects merged into one component, yields a single averaged motion for distinct objects, while over-segmentation fragments the flow of a single object.
Both failure modes are governed by the pixel size $s$, the dilation, and the connectivity, and although the validation step of Sec.~\ref{sec:clustering} mitigates the former, neither is fully eliminated.
Third, both variants estimate translation only: the dense variant assigns one displacement per component and the key-point variant one median displacement, so intra-frame rotation and articulation are not captured; this is benign at typical frame rates but degrades for long temporal baselines, where the constant-velocity assumption of Sec.~\ref{sec:leverage_past_pcs} also weakens for accelerating or turning objects.
Fourth, the key-point variant additionally depends on the presence of repeatable boundary structure and on the gating radius; featureless or heavily occluded footprints yield few confident matches, and the corresponding components conservatively receive zero flow.
Finally, like all classical pipelines, the method presumes accurate ego-motion compensation; pose errors translate directly into spurious flow, particularly for distant points where small rotational errors cause large displacements.

\section{Conclusion}

We presented CorrelationFlow, a training-free family of LiDAR scene flow estimators that reduces the task to two classical operations: image correlation and connected-component labeling on a bird's-eye-view occupancy projection.
The dense variant recovers per-component motion by correlation maximization over a temporal window, robustified by a median over multiple baselines; the key-point variant, CorrelationFlow-Keypoints, operates on a single sweep pair by matching boundary key points derived from the distance transform, filtered by Lowe's ratio test, and propagating their flow within connected components.
Neither variant requires training data, labels, or learned parameters and both expose a single interpretable correlation operation at their core.
In the Argoverse 2 2026 Multi-Dataset Scene Flow Challenge, CorrelationFlow ranks second on the unsupervised track, demonstrating that classical geometric operations remain competitive with learned methods across heterogeneous sensors and platforms.
Consistent with recent re-evaluations of LiDAR scene flow \cite{khatri2024trackflow}, we argue that such transparent baselines are valuable both as practical, reproducible estimators and as a yardstick for how much of the problem is solvable without learning.
Future work includes lifting the ground-plane restriction through multi-height occupancy slices, modeling per-component rotation, and coupling the two variants more tightly, e.g., by using key-point matches to seed or verify the dense correlation.

\section*{Acknowledgment} 

We would like to acknowledge the use of ChatGPT by OpenAI / Claude by Anthropic for proofreading and grammar checking during the preparation of this manuscript. The tool was used to suggest corrections for spelling, grammar, and sentence flow. The final responsibility for the content, accuracy, and integrity of this work rests entirely with the author(s).




\end{document}